\documentclass[acmtkdd]{acmtrans2m}

\acmVolume{x}
\acmNumber{y}
\acmYear{10}
\acmMonth{zz}

\usepackage{times}
\usepackage{graphicx}

\newcommand{\BibTeX}{{\rm B\kern-.05em{\sc i\kern-.025em b}\kern-.08em
    T\kern-.1667em\lower.7ex\hbox{E}\kern-.125emX}}
\newcommand{\secref}[1]{Section~\ref{#1}}
\newcommand{\eqref}[1]{Eq.~(\ref{#1})}
\newcommand{\figref}[1]{Figure~\ref{#1}}
\newcommand{\tabref}[1]{Table~\ref{#1}}

\newcommand{\commentout}[1]{}
\newcommand{\comment}[1]{}

\title{Modeling Social Annotation: a Bayesian Approach}

\author{Anon Plangprasopchok and Kristina Lerman \\ USC Information Sciences Institute}

\begin{abstract}
Collaborative tagging systems, such as \emph{Delicious}, \emph{CiteULike}, and others, allow users to annotate resources, e.g., Web pages or scientific papers, with descriptive labels called tags.
The social annotations contributed by thousands of users, can potentially be used to infer categorical knowledge, classify documents or recommend new relevant information. Traditional text inference methods do not make best use of social annotation, since they do not take into account variations  in individual users' perspectives and vocabulary.
In a previous work, we introduced a simple probabilistic model that takes interests of individual annotators into account in order to find hidden topics of annotated resources.
Unfortunately, that approach had one major shortcoming: the number of topics and interests must be specified \emph{a priori}.
To address this drawback, we extend the model to a fully Bayesian framework, which offers a way to automatically estimate these numbers. In particular, the model allows the number of interests and topics to change as suggested by the structure of the data.
We evaluate the proposed model in detail on the synthetic and real-world data by comparing its performance to Latent Dirichlet Allocation on the topic extraction task.
For the latter evaluation, we apply the model to infer topics of Web resources from social annotations obtained from \emph{Delicious} in order to discover new resources similar to a specified one. Our empirical results demonstrate that the proposed model is a promising method for exploiting social knowledge contained in user-generated annotations.
\end{abstract}


\begin{document}

\setcounter{page}{1}
\begin{bottomstuff}
Author's address: USC Information Sciences Institute, 4676 Admiralty Way, Marina del Rey, CA 90292.
\end{bottomstuff}

\maketitle

\section{Introduction}
A new generation of Social Web sites, such as \emph{Delicious}, \emph{Flickr}, \emph{CiteULike}, \emph{YouTube}, and others, allow users to share content and annotate it with metadata in the form of comments, notes, ratings, and descriptive labels known as tags. Social annotation captures the collective knowledge of thousands of users and can potentially be used to enhance a number of applications including Web search, information personalization and recommendation, and even synthesize categorical knowledge~\cite{SchmitzTagging06,Mika07_OntoAreUs}. In order to make best use of user-generated metadata, we need methods that effectively deal with the challenges of data sparseness and noise, as well as take into account variations in the vocabulary, interests, and the level of expertise among individual users.

Consider specifically \emph{tagging}, which has become a popular method for annotating content on the Social Web. When a user tags a resource, be it a Web page on the social bookmarking cite \emph{Delicious}, a scientific paper on \emph{CiteULike}, or an image on the social photosharing site \emph{Flickr}, the user is free to select any keyword, or tag, from an uncontrolled personal vocabulary to describe the resource. We can use tags to categorize resources, similar to the way documents are categorized using their text, although the usual problems of sparseness (few unique keywords per document), synonymy (different keywords may have the same meaning), and ambiguity (same keyword has multiple meanings), will also be present in this domain. Dimensionality reduction techniques such as topic modeling~\cite{HofmannPLSA99,BleiLDA03,BuntinePCA04}, which project documents from word space to a dense topic space, can alleviate these problems to a certain degree.
%[feb22] try to mention that we already know (have enough knowledge on) the utility of existing topic modeling approaches which they can resolve sparseness, polysemy and synonymy problems.
Specifically, such projections address the sparseness and synonymy challenges by combining ``similar'' words together in a topic. Similarly, the challenge of word ambiguity in a document is addressed by taking into account the senses of co-appearing words in that document. In other words, the sense of the word is determined jointly along with the other words in that document.

Straightforward application of the previously mentioned methods to social annotation would aggregate resource's tags over all users, thereby losing important information about individual variation in tag usage, which can actually help the categorization task. Specifically, in social annotation, similar tags may have different meanings according to annotators' perspectives on the resource they are annotating \cite{Lerman07flickrsearch}. For example, if one searches for Web resources about car prices using the tag ``jaguar'' on \emph{Delicious}, one receives back a list of resources containing documents about luxury cars and dealers, as well as guitar manuals, wildlife videos, and documents about Apple Computer's operating system. The above mentioned methods would simply compute tag occurrences from annotations across all users, effectively treating all annotations as if they were coming from a single user. As a result, a resource annotated with the tag ``jaguar'' will be undesirably associated with any sense of the keyword simply based on the number of times that keyword (tag) was used for each sense.

% contributions of the paper
We claim that users express their individual interests and vocabulary through tags, and that we can use this information to  learn a better topic model of tagged resources. For instance, we are likely to discover that users who are interested in luxury cars use the keyword ``jaguar'' to tag car-related Web pages\comment{and those interested in vintage cars use ``jaguar'' to tag vintage car-related Web pages}; while, those who are interested in wildlife use ``jaguar'' to tag wildlife-related Web pages.
The additional information about user interests is essential, especially since social annotations are generally very sparse.\footnote{There are only $3.74$ tags on average for a certain photo in Flickr \cite{Rattenbury07}. In addition, there are $4$ to $7$ tags used by each user on a certain URL from our observation in \emph{Delicious} data we obtained; while tag vocabulary on an resource gets stable after few bookmarks as reported in \cite{Golder06}.}
In a previous work, \cite{delicious07::iiweb}, we proposed a probabilistic model that takes into account interest variation among users to infer a more accurate topic model of tagged resources. In this paper  we describe a Bayesian version of the model (\secref{sec:finiteitm}). We explore its performance in detail on the synthetic data (\secref{sec:synthetic}) and compare it to Latent Dirichlet Allocation (LDA)~\cite{BleiLDA03}, a popular document modeling algorithm.
We show that in domains with high tag ambiguity, variations among users can actually help discriminate between tag senses, leading to a better topic model. Our model is, therefore, best suited to make sense of social metadata, since this domain is characterized both by a high degree of noise and ambiguity and a highly diverse user population with varied interests.

As a second contribution of the paper, we incorporate a Hierarchical Dirichlet Process~\cite{Teh04hierarchicaldirichlet} to create an adaptive version of the proposed model (\secref{sec:infiniteitm}), which enables the learning method to automatically adjust the \comment{dimensions of }model parameters. This capability helps overcome one of the main difficulties of applying the original model to the data: namely, having to specify the right number of topics and interests.

% AP 12/05/09
Finally, the proposed models are validated on a real-world data set obtained from the social bookmarking site \emph{Delicious} (\secref{sec:realitm} and \secref{sec:nonparamrealeval}). We first train the model on this data, then measure the quality of the learned topic model. Specifically, the learned topic model is used as a compressed description of each Web resource. We compute similarity between resources based on the compressed description and manually evaluate results to show that the topic model obtained by the method proposed in this paper identifies more similar resources than the baseline.

\section{Modeling Social Annotation}
\label{sec:definition}

In general, a social annotation system involves three entities: resources (e.g., Web pages on \emph{Delicious}), users and metadata. Although there are different forms of metadata, such as descriptions, notes and ratings, we focus on tags only in this context. We define a variable $R$ as resources, $U$ as users, and $T$ as tags. Their realizations are defined as $r$, $u$ and $t$ respectively. A post (or bookmark) $k$ on a resource $r$ by a user $u$, can be formalized as a tuple $\langle r,u,\{t_{1},t_{2},\ldots ,t_{j}\}\rangle_{k}$, which can be further broken down into co-occurrence of $j$ resource-user-tag triples: $\langle r,u,t \rangle$. \comment{Hence, there will be $j$ triples for this bookmark.} $N_{R}$, $N_{U}$ and $N_{T}$ are the number of distinct resources, users and tags respectively.

In addition to the observable variables defined above, we introduce two `hidden' or `latent' variables, which we will attempt to infer from the observed data.
The first variable, $Z$, represents resource topics, which we view as categories or concepts of resources. From our previous example, the tag ``jaguar'' can be associated with topics `cars', `animals', `South America', `computers', etc. The second variable, $X$,  represents user interests, the degree to which users subscribe to these concepts.
One user may be interested in collecting information about luxury cars before purchasing one, while another user may be interested in vintage cars.
A user $u$ has her interest profile, $\psi_{u}$, which is a weight distribution over all possible interests $x$. And $\psi$ (without subscript) is just  an $N_{U} \times N_{X}$ matrix. Similarly, a resource $r$ has its topic profile, $\phi_{r}$, which is again a weight distribution over all possible topics $z$, whereas $\phi$ (without subscript) is an $N_{R} \times N_{Z}$ matrix. Thus, a resource about South American jaguars will have a higher weight on `animals' and `South America' topics than on the `cars' topic.
Usage of tags for a certain interest-topic pair $(x,z)$ is defined as a weight distribution over tags, $\theta_{x,z}$ -- that is, some tags are more likely to occur for a given pair than others. The weight distribution of all tags, $\theta$, can be viewed as an $N_{T} \times N_{Z} \times N_{X}$ matrix.

\begin{figure}
\begin{center}
\begin{tabular}{cc}
\includegraphics[width=2.6in]{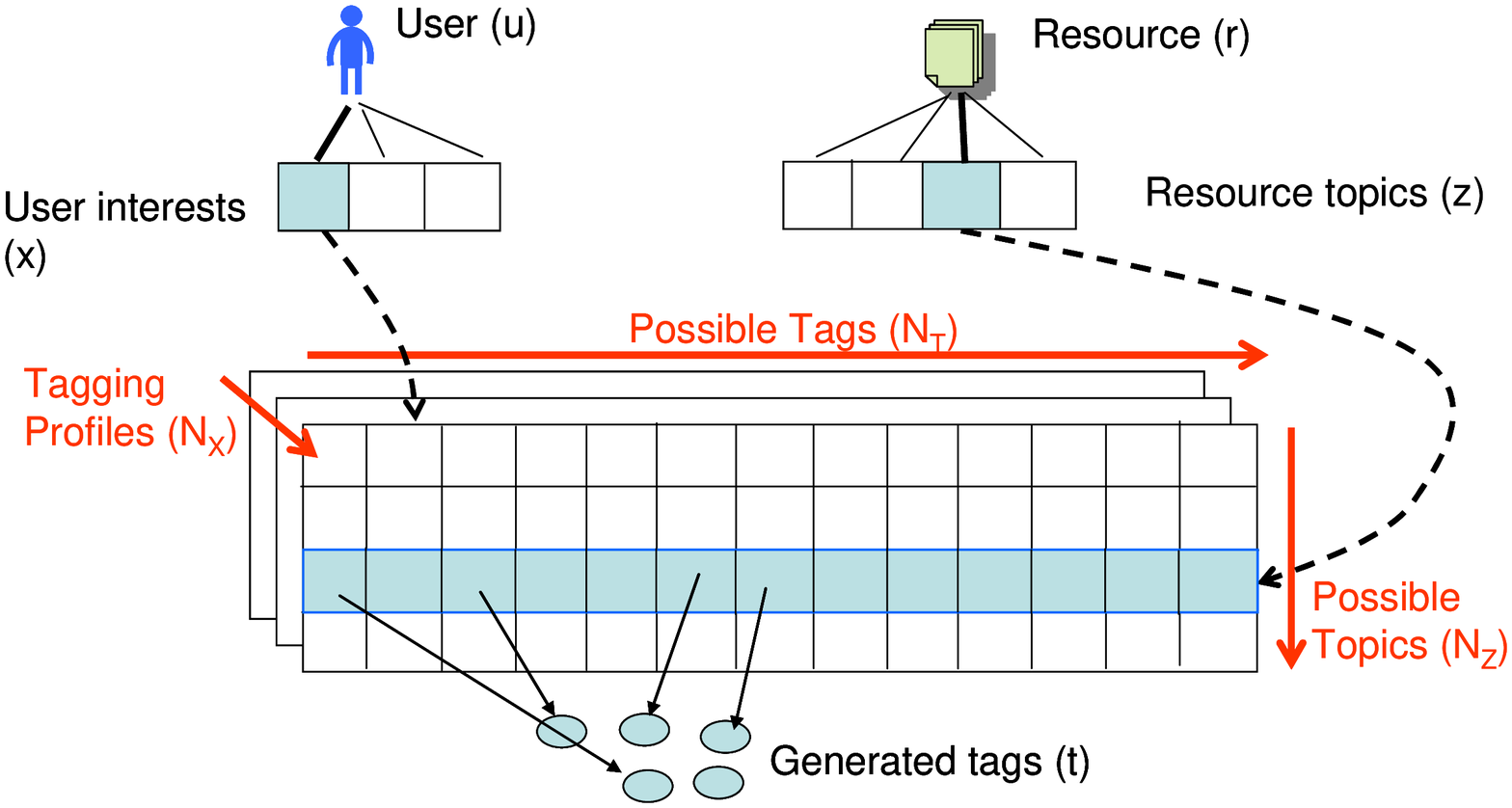} &
\includegraphics[width=2.2in]{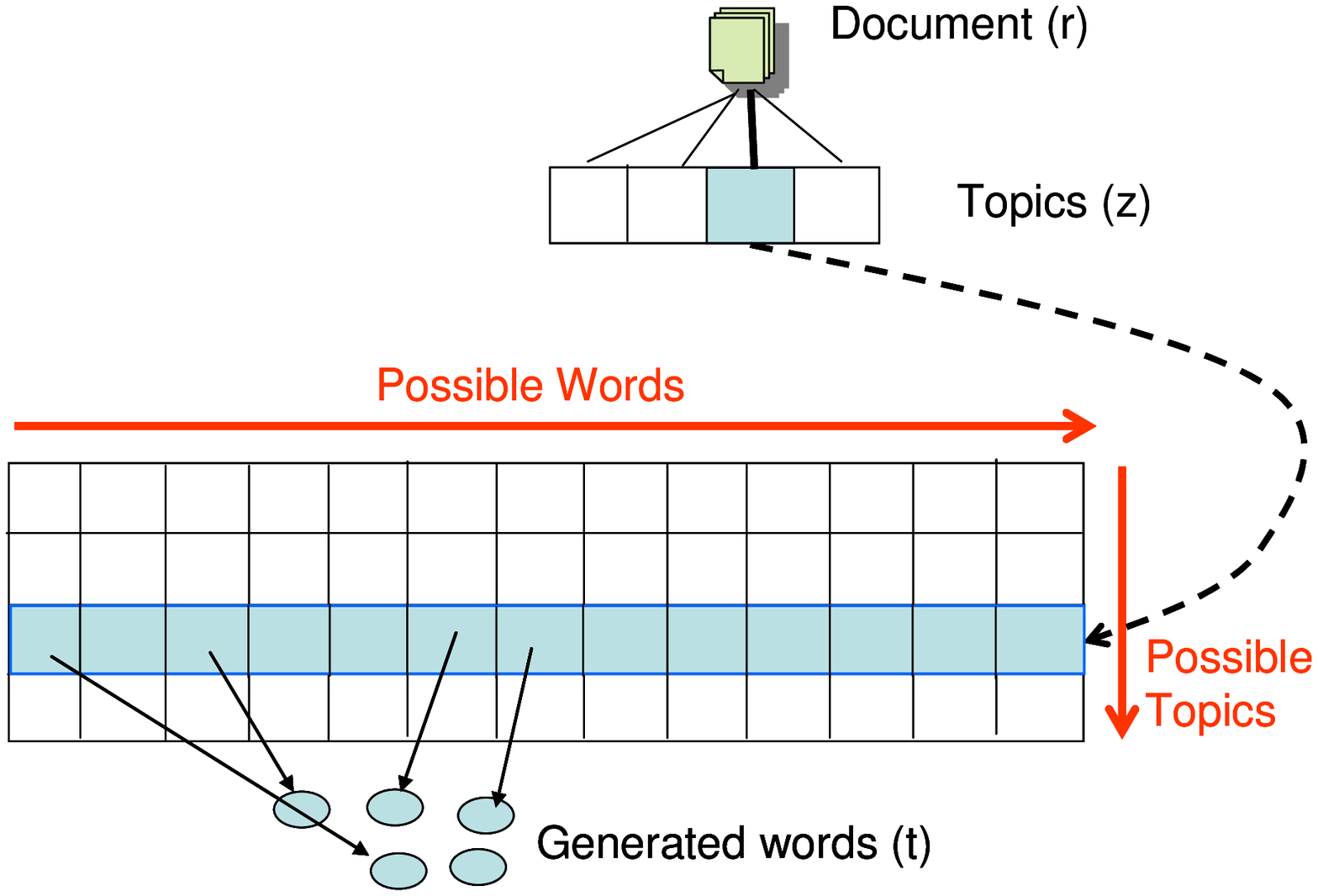} \\
(a) Social Annotation Process & (b) Document Word Generation Process
\end{tabular}
\end{center}
\caption {Schematic diagrams represent: (a) tag generation process in social annotation domain; (b) word generation process in document modeling domain.
} \label{fig:schematic}
\end{figure}

We cast an annotation event as a stochastic process as follows:
\begin{itemize}
\item User $u$ finds a resource $r$ interesting and would like to bookmark it.

\item For each tag that $u$ generates for $r$:

\begin{itemize}
\item User $u$ selects an interest $x$ from her interest profile $\psi_{u}$; resource $r$ selects a topic $z$ from its topic profile $\phi_{r}$.

\item Tag $t$ is then chosen based on users's interest and resource's topic from the interest-topic distribution over all tags $\theta_{x,z}$.
\end{itemize}

\end{itemize}
\noindent This process is depicted schematically in \figref{fig:schematic}(a). Specifically, a user $u$ has an interest profile, represented by a vector of interests $\psi_{u}$. Meanwhile, a resource $r$ has its own topic profile, represented by a vector of topics $\phi_{r}$. Users who share the same interest ($x$) have the same tagging policy --- the tagging profile ``plate'', \comment{$\theta_{x}$, as}shown in the diagram. For the ``plate'' corresponding to an interest $x$, each row corresponds to a particular topic $z$, and it gives $\theta_{x,z}$, the distribution over all tags for that topic and interest.

%[feb22] comparison to standard topic modeling.
The process can be compared to the word generation process in standard topic modeling approaches, e.g., LDA~\cite{BleiLDA03} and pLSA~\cite{HofmannPLSA2001}, as shown in \figref{fig:schematic} (b). In topic modeling, words of a certain document are generated according to a single policy, which assumes that all authors of documents in the corpus share the same tagging patterns. In other words, a set of ``similar'' tags is used to represent a topic across all authors.
% KL 12/10
In our ``jaguar'' example, for instance, we may find one topic to be strongly associated with words ``cars'', ``automotive'', ``parts'', ``jag'', etc., while another topic may be associated with words ``animals'', ``cats'', ``cute'', ``black'', etc., and still another with ``guitar'', ``fender'', ``music'', etc. and so on.

In social annotation, however, a resource can be annotated by many users, who may have different opinions, even on the same topic. Users who are interested in restoring vintage cars will have a different tagging profile than those who are interested in shopping for a luxury car.
% KL 12/10
The `cars' topic would then decompose under different tagging profiles into one that is highly associated with words ``restoration'', ``classic'', ``parts'', ``catalog'', etc., and another that is associated with words ``luxury'', ``design'', ``performance'', ``brand'', etc.
The separation of tagging profiles for each group of users who share the same interest provides a machinery to address this issue and constitutes the major distinction between our work and standard topic modeling.

\section{Finite Interest Topic Model}
\label{sec:finiteitm}
In our previous work \cite{delicious07::iiweb}, we proposed a probabilistic model that describes social annotation process, which was extended from probabilistic Latent Semantic Analysis (pLSA) \cite{HofmannPLSA2001}.
However, the model inherited some shortcomings from pLSA. First, the strategy for estimating parameters in both models --- the point estimation using EM algorithm --- has been criticized as being prone to local maxima \cite{GriffithsTopic04,TopicModel06}. In addition, there is also no explicit way to extend these models to automatically infer the dimensions of parameters, such as the number of components used to represent topics ($N_{Z}$) and interests ($N_{X}$).

\begin{figure}
\begin{center}
\includegraphics[width=3.1in]{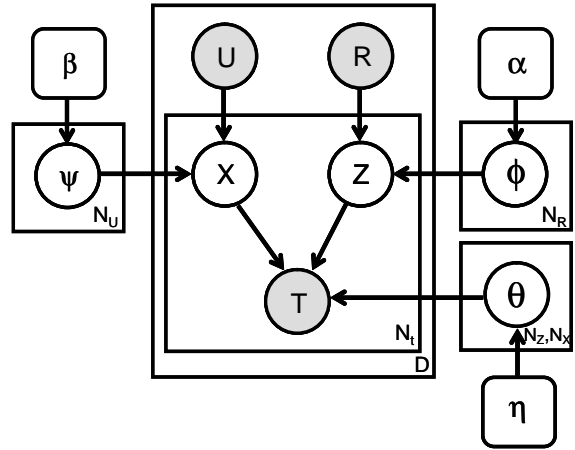}
\end{center}
\caption {Graphical representation of the social annotation process. $R$, $U$,
$T$, $X$ and $Z$ denote variables ``Resource'', ``User'', ``Tag'',
``Interest'' and  ``Topic'' respectively. $\psi$, $\phi$ and $\theta$
are distributions of user over interests, resource over topics and
interest-topic over tags respectively. $N_{t}$ represents the number
of tag occurrences for one bookmark (by a particular user, $u$, on a
particular resource, $r$); $D$ represents the number of all bookmarks in
the social annotation system. The hyperparameters $\alpha$, $\beta$, and $\eta$ control dispersions of categorical topics, interests and tags respectively.
 } \label{fig:bitm}
\end{figure}

We extend our previous Interest Topic Model (ITM) the same way pLSA was upgraded to Latent Dirichlet Allocation (LDA) model~\cite{BleiLDA03}. In other words, we implement the model under a Bayesian framework, which offers solutions \cite{BleiLDA03,GriffithsTopic04,NealDPM00} to the previously mentioned problems. By doing so, we introduce priors on top of parameters $\psi$, $\phi$ and $\theta$ to make the model fully generative, i.e., the mechanism for generating these parameters is explicitly implemented.
\commentout{As the strong assumption that numbers of topics $N_{Z}$ and interests $N_{X}$ must be fixed and known a priori, is still hold. And we assume that most users share some similar interests; as most resources share some similar topics. Thus, we expect $N_{X} \ll N_{U}$ and $N_{Z} \ll N_{R}$.}
To make the model analytically simple, we use symmetric Dirichlet priors. Following the generative process described in \secref{sec:definition}, the model can be described as a stochastic process, depicted a graphical form~\cite{Buntine94} in \figref{fig:bitm}:
\begin{description}
\item $\psi_{u} \sim Dirichlet(\beta/N_{X},...,\beta/N_{X})$ (generating user $u$ interest's profile)

\item $\phi_{r} \sim Dirichlet(\alpha/N_{Z},...,\alpha/N_{Z})$ (generating resource $r$ topic's profile)

\item $\theta_{x,z} \sim Dirichlet(\eta/N_{T},...,\eta/N_{T})$. (generating tag's profile for interest $x$ and topic $z$)
\end{description}

\noindent For each tag $t_{i}$ of a bookmark, 
\begin{description}
    \item $x_{i} \sim Discrete(\psi_{u})$
  \item $z_{i} \sim Discrete(\phi_{r})$
  \item $t_{i} \sim Discrete(\theta_{x_{i},z_{i}})$.
\end{description}

One possible way to estimate parameters is to use Gibbs sampling \cite{Gil96,NealDPM00}. Briefly, the idea behind the Gibbs sampling is to iteratively use the parameters of the current state to estimate parameters of the next state. In particular, each next-state parameter is sampled from the posterior distribution of that parameter given all other parameters in the previous state. The sampling process is done sequentially until sampled parameters approach the target posterior distributions. Recently, this approach was demonstrated to be simple to implement, yet competitively efficient, and to yield relatively good performance on the topic extraction task \cite{GriffithsTopic04,TopicModelSmyth2004}.

Since we use Dirichlet priors, it is straightforward to integrate out $\psi$, $\phi$ and $\theta$. Thus, we only need to sample hidden variables $\textbf{x}$ and $\textbf{z}$ and later on estimate $\psi$, $\phi$ and $\theta$ once $\textbf{x}$ and $\textbf{z}$ approach their target posterior distribution. To derive Gibbs sampling formula for sampling $\textbf{x}$ and $\textbf{z}$, we first assume that all bookmarks are broken into $N_{K}$ tuples. Each tuple is indexed by $i$ and we refer to the observable variables, resource, user and tag, of the tuple $i$ as $r_{i}$, $u_{i}$, $t_{i}$. We refer to the hidden variables, topic and interest, for this tuple as $z_{i}$ and $x_{i}$ respectively, with $\textbf{x}$ and $\textbf{z}$ representing the vector of interests and topics over all tuples.

% AP120509: modified a little bit here
% In words, if $z = z_{i}$, $N_{r_{i},z_{-i}} = N_{r_{i},z_{i}} -1 $; otherwise, $N_{r_{i},z_{-i}} = N_{r_{i},z_{i}}$.
We define $N_{r_{i},z_{-i}}$ as the number of all tuples having $r = r_{i}$ and $z$ but excluding the present tuple $i$. In words, if $z = z_{i}$, $N_{r_{i},z_{-i}} = N_{r_{i},z_{i}}-1$; otherwise, $N_{r_{i},z_{-i}} = N_{r_{i},z_{i}}$. Similarly, $N_{z_{-i},x_{i},t_{i}}$ is a number of all tuples having $x = x_{i}$, $t = t_{i}$ and $z$ but excluding the present tuple $i$; $\textbf{z}_{-i}$ represents all topic assignments except that of the tuple $i$. The Gibbs sampling formulas for sampling $z$ and $x$, whose derivation we provide in the Appendix, are as follows.

\begin{eqnarray}
p(z_{i}|\textbf{z}_{-i},\textbf{x},\textbf{t}) = \frac{N_{r_{i},z_{-i}}+\alpha/N_{Z}}{N_{r_{i}}+\alpha-1}.\frac{N_{z_{-i},x_{i},t_{i}}+\eta/N_{T}}{N_{z_{-i},x_{i}}+\eta}
\label{eq:gibbsZ}
\end{eqnarray}

\begin{eqnarray}
p(x_{i}|\textbf{x}_{-i},\textbf{z},\textbf{t}) = \frac{N_{u_{i},x_{-i}}+\beta/N_{X}}{N_{u_{i}}+\beta-1}\cdot\frac{N_{x_{-i},z_{i},t_{i}}+\eta/N_{T}}{N_{x_{-i},z_{i}}+\eta}
\label{eq:gibbsX}
\end{eqnarray}

%[modified feb23]: semantic on eq:gibbsZ and eq:gibbsX
Consider \eqref{eq:gibbsZ}, which computes a probability of a certain topic for the present tuple. This equation is composed of 2 factors. Suppose that we are currently determining the probability that the topic of the present tuple $i$ is $j$ ($z_{i} = j$). The left factor determines the probability of topic $j$ to which the resource $r_{i}$ belongs according to the present topic distribution \comment{of the other tags} of $r_{i}$. Meanwhile, the right factor determines the probability of tag $t_{i}$ under the topic $j$ of the users who have interest $x_{i}$. If resource $r_{i}$ assigned to the topic $j$  has many tags, and the present tag $t_{i}$ is ``very important'' to the topic $j$ according to the users with interest $x_{i}$, there is a higher chance that tuple $i$ will be assigned to topic $j$. A similar insight is also applicable to \eqref{eq:gibbsX}. In particular, suppose that we are currently determining the probability that the interest of the present tuple $i$ is $k$ ($x_{i} = k$). If user $u_{i}$ assigned to the interest $k$ has many tags, and tag $t_{i}$ is ``very important'' to the topic $z_{i}$ according to users with interest $k$, the tuple $i$ will be assigned to interest $k$ with higher probability.

In the model training process,  we sample topic $z$ and interest $x$ in the current iteration using their assignments from the previous iteration. By sampling $z$ and $x$ using \eqref{eq:gibbsZ} and \eqref{eq:gibbsX} for each tuple, the posterior distribution of topics and interests is expected to converge to the true posterior distribution after enough iterations. Although it is  difficult to assess convergence of Gibbs sampler in some cases as mentioned in \cite{Sahu99onconvergence}, we simply monitor it through the likelihood of data given the model, which measures how well the estimated parameters fit to the data.  Once the likelihood reaches the stable state, it only slightly fluctuates from one iteration to the next, i.e., there is no systematic and significant increase and decrease in likelihood. We can use this as a part of the stopping criterion. Specifically, we monitor likelihood changes over a number of consecutive iterations. If the average of these changes is less than some threshold, the estimation process terminates. More robust approaches to determining the stable state are discussed elsewhere, e.g. \cite{TannerGibbsStopper}. The formula for the likelihood is defined as follows.

\begin{eqnarray}
f(\textbf{t};\psi,\phi,\theta) = \prod_{i=1:N_{K}}{ \big( \frac{N_{x_{i},z_{i},t_{i}}+\eta/N_{T}}{N_{x_{i},z_{i}}+\eta} \big) }
\label{eq:likelihood}
\end{eqnarray}

To avoid a numerical precision problem in model implementation, one usually uses log likelihood $log(f(\textbf{t};\psi,\phi,\theta))$ instead. Note that we use the strategy mentioned in \cite{Escobar95bayesiandensity} (Section 6) to estimate $\alpha$, $\beta$ and $\eta$ from data.

The sampling results in the stable state are used to estimate model parameters. Again, we define $N_{r,z}$ as the number of all tuples associated with resource $r$ and topic $z$, with $N_{r}$, $N_{x,u}$, $N_{u}$, $N_{x,u,t}$ and $N_{x,z}$ defined in a similar way. From \eqref{eq:posteriorz} and \eqref{eq:posteriorx} in the Appendix, the formulas for computing such parameters are as follows:

\begin{eqnarray}
\phi_{r,z} = \frac{N_{r,z}+\alpha/N_{Z}}{N_{r}+\alpha}
\label{eq:phi}
\end{eqnarray}

\begin{eqnarray}
\psi_{u,x} = \frac{N_{u,x}+\beta/N_{X}}{N_{u}+\beta}
\label{eq:psi}
\end{eqnarray}

\begin{eqnarray}
\theta_{x,z,t} = \frac{N_{x,z,t}+\eta/N_{T}}{N_{x,z}+\eta}
\label{eq:theta}
\end{eqnarray}

% add something about Gibbs at stable state...
Parameter estimation via Gibbs sampling is less prone to the local maxima problem than the generic EM algorithm, as argued in \cite{TopicModelSmyth2004}. In particular, this scheme does not estimate parameters $\phi$, $\psi$, and $\theta$ directly. Rather, they are integrated out, while the hidden variables $z$ and $x$ are iteratively sampled during the training process. The process estimates the ``posterior distribution'' over possible values of $\phi$, $\psi$, and $\theta$. At a stable state, $z$ and $x$ are drawn from this distribution and then used to estimate $\phi$, $\psi$, and $\theta$. Consequently, these parameters are estimated from a combination of ``most probable solutions'', which are obtained from multiple maxima. This clearly differs from the generic EM with point estimation, which we used in our previous work \cite{delicious07::iiweb}. Specifically, the point estimation scheme estimates $\phi$, $\psi$, and $\theta$ from single local maximum.

Per training iteration, the computational complexity of Gibbs sampling is more expensive than EM. This is because we need to sample hidden variables ($z$ and  $x$) for each data point (tuple), whereas EM only requires updating parameters. In general, the number of the data points is larger than the dimension of parameters. However, it has been reported in \cite{GriffithsTopic04} that to reach the same performance, Gibbs sampling requires fewer floating point operations than the other popular approaches: Variational Bayes and Expectation Propagation \cite{MinkaEP01}. Moreover, to our knowledge, there is currently no explicit way to extend these approaches to automatically infer the size of hidden variables, as Gibbs sampling can. Note that inference of these numbers is described in \secref{sec:infiniteitm}.

\section{Evaluation}
\label{sec:eval}
In this section we evaluate the Interest Topic Model and compare its performance to LDA~\cite{BleiLDA03} on both synthetic and real-world data. The synthetic data set enables us to control the degree of tag ambiguity and individual user variation, and examine in detail how both learning algorithms respond to these key challenges of learning from social metadata. The real-world data set, obtained from the social bookmarking site \emph{Delicious}, demonstrates the utility of the proposed model. 

The baseline in both comparisons is LDA, a probabilistic generative model originally developed for modeling text documents \cite{BleiLDA03}, and more recently extended to other domains, such as finding topics of scientific papers \cite{GriffithsTopic04}, topic-author associations \cite{TopicModelSmyth2004}, user roles in a social network \cite{McCallumWC07}, and Collaborative Filtering \cite{BenMarlin:2004}.  In this model, the distribution of a document over a set of topics is first sampled from a Dirichlet prior. For generating each word in the document, a topic is first sampled from the distribution; then, a word is selected from the distribution of topics over words. One can apply LDA to model how tags are generated for resources on social tagging systems. One straightforward approach is to ignore information about users, treating all tags as if they came from the same user. Then, a resource can be viewed as a document, while tags across different users who bookmarked it are treated as words, and LDA is then used to learn parameters.

ITM extends LDA by taking into account individual variations among users. In particular, a tag for a certain bookmark is chosen not only from the resource's topics but also from user's interests. This allows each user group (with the same interest) to have its own policy, $\theta_{x,z,t}$, for choosing tags to represent a topic. Each policy is then used to update resource topics as in \eqref{eq:gibbsZ}.  Consequently, $\phi_{r,z}$ is updated based on interests of users who actually annotated resource $r$, rather than updating it from a single policy that ignores user information.
We thus expect ITM to perform better than LDA when annotations are made by diverse user groups, and especially when tags are ambiguous.

\subsection{Synthetic Data}
\label{sec:synthetic}

To verify the intuition about ITM, we evaluated the performance of the learning algorithms on synthetic data. Our data set consists of 40 resources, 10 topics, 100 users, 10 interests, and 100 tags.
We first separate resources into five groups, with resources in each group assigned topic weights from the same (Dirichlet) probability distribution, which forces each resource to favor 2--4 out of ten topics.
Rather than simulate the tagging behavior of user groups by generating individual tagging policy plates as in \figref{fig:schematic}(a), we simplify the generative process to simulate the impact of diversity in user interests on tagging. To this end, we represent user interests as distributions over topics.

We create data sets under different tag ambiguity and user interest variation levels. To make these settings \emph{tunable}, we generate distributions of topics over tags, and distributions of resources over topics using symmetric Dirichlet distributions with different parameter values. Intuitively, when sampling from the symmetric Dirichlet distribution\footnote{Samples that are sampled from Dirichlet distribution are discrete probability distributions} with a low parameter value, for example 0.01, the sampled distribution contributes weights (probability values that are greater than zero) to only a few elements. In contrast, the distribution will contribute weights to many elements when it is sampled from a Dirichlet distribution with a high parameter value.  We used this parameter of the symmetric Dirichlet distribution to adjust  \emph{user variation}\comment{($\beta$)}, i.e., how broad or narrow user interests are, and \emph{tag ambiguity} \comment{($\epsilon$)}, i.e., how many or how few topics each tag belongs to. With higher parameter values, we can simulate a behavior of more ambiguous tags, such as ``jaguar'', which has multiple senses, i.e., it has weights allocated to many topics. Low parameter values can be used to simulate low ambiguity tags, such as ``mammal'', which has one or few senses. The parameter values used in the experiments are 1, 0.5, 0.1, 0.05 and 0.01.

To generate tags for each simulated data set, user interest profiles $\psi_{u}$ are first drawn from the symmetric Dirichlet distribution with the same parameter value. A similar procedure is done for distributions of topics over words $\theta$. A resource will presumably be annotated by a user if the match between resource's topics and user's interests is greater than some threshold. The match is given by the inner product between the resource's topics and user's interests, and we set the threshold at $1.5 \times$ the average match computed over all user-resource pairs.
The rationale behind this choice of threshold is to ensure that a resource will be tagged by a user who is strongly interested in the topics of that resource. When the user-resource match is greater than threshold, a set of tags (a post or bookmark) is generated according to the following procedure. First, we compute the topic distribution from an element-wise product of the resource's topics and user's interests. Next, we sample a topic from this distribution and produce a tag from the tag distribution of that topic. This guarantees that tags are only generated according to user's interests. We repeat this process seven times in each post\footnote{We chose seven because \emph{Delicious} users in general use four to seven tags in each post.} and eliminate redundant tags.
The process of generating tags is summarized below:
\comment{
\begin{algorithmic}
\FOR{each resource-user pair ($u$,$r$)}
\STATE $m_{r,u} = \phi_{r} \cdot \psi_{u}$ (compute the match score)
\ENDFOR
\STATE $\bar{m} = Average(\textbf{m})$
\FOR{each resource-user pair ($r$,$u$)}
\IF{$m_{r,u} > 1.5  \bar{m}$ }
\STATE $topicPref = \phi_{r} \times \psi_{u}$ (element-wise product)
\FOR{$i = 1$ to $7$}
\STATE $z \sim topicPref$ (draw a topic from the topic preference)
\STATE $t_{r,u}^{i} \sim \theta_{z}$ (sample $i^{th}$ tag for the ($u$,$r$) pair)
\ENDFOR
\STATE Remove redundant tags
\ENDIF
\ENDFOR
\end{algorithmic}
}
\begin{table}[tbh*]
\begin{tabular}{llll}
\multicolumn{4}{l}{\textbf{for} each resource-user pair ($u$,$r$) \textbf{do}} \\ 
 & \multicolumn{3}{l}{$m_{r,u} = \phi_{r} \cdot \psi_{u}$ (compute the match score)} \\ 
\multicolumn{4}{l}{\textbf{end for}} \\ 
\multicolumn{4}{l}{$\bar{m} = Average(\textbf{m})$} \\ 
\multicolumn{4}{l}{\textbf{for} each resource-user pair ($r$,$u$) \textbf{do}} \\ 
 & \multicolumn{3}{l}{\textbf{if} $m_{r,u} > 1.5  \bar{m}$ \textbf{then}} \\ 
 &  & \multicolumn{2}{l}{$topicPref = \phi_{r} \times \psi_{u}$ (element-wise product)} \\ 
 &  & \multicolumn{2}{l}{\textbf{for} $i = 1$ to $7$ \textbf{do}} \\ 
 &  &  & $z \sim topicPref$ (draw a topic from the topic preference) \\ 
 &  &  & $t_{r,u}^{i} \sim \theta_{z}$ (sample $i^{th}$ tag for the ($u$,$r$) pair) \\ 
 &  & \multicolumn{2}{l}{\textbf{end for}} \\ 
 &  & \multicolumn{2}{l}{Remove redundant tags} \\ 
 & \multicolumn{3}{l}{\textbf{end if}} \\ 
\multicolumn{4}{l}{\textbf{end for}} \\ 
\end{tabular}
\end{table}

We measure sensitivity to tag ambiguity and user interest variation for LDA and ITM on the synthetic data generated with different values of symmetric Dirichlet parameters. One way to measure sensitivity is to determine how the learned topic distribution, $\phi_{r}^{ITM}$ or $\phi_{r}^{LDA}$, deviates from the actual topic distribution of resource $r$, $\phi_{r}^{Actual}$. Unfortunately, we cannot compare them directly, since topic order of the learned topic distribution may not be the same as that of the actual one.\footnote{This property of probabilistic topic models is called exchangeability of topics~\protect\cite{TopicModel06}.} An indirect way to measure this deviation is to compare distances between pairs of resources computed using the actual and learned topic distributions. We define this deviation as $\Delta$. We calculate the distance between two distributions using Jensen-Shannon divergence (JSD)~\cite{JSDiverge}. If a model accurately learned the resources' topic distribution, the distance between two resources computed using the learned distribution will be equal to the distance computed from the actual distribution. Hence, the lower $\Delta$, the better model performance.
The deviation between the actual and learned topic distributions is
\begin{eqnarray}
\Delta = \sum_{r=1}^{N_{R}}{\sum_{r^{\prime}=r+1}^{N_{R}}{|JSD(\phi_{r}^{Learned},\phi_{r^{\prime}}^{Learned}) - JSD(\phi_{r}^{Actual},\phi_{r^{\prime}}^{Actual})|}}.
\label{eq:avgdivITM}
\end{eqnarray}
\noindent $\Delta$  is computed separately for each algorithm, ${Learned}=ITM$ and $Learned=LDA$.

% put some simulation result here
\begin{figure}
\begin{center}
\begin{tabular}{cc}
\includegraphics[width=2.2in]{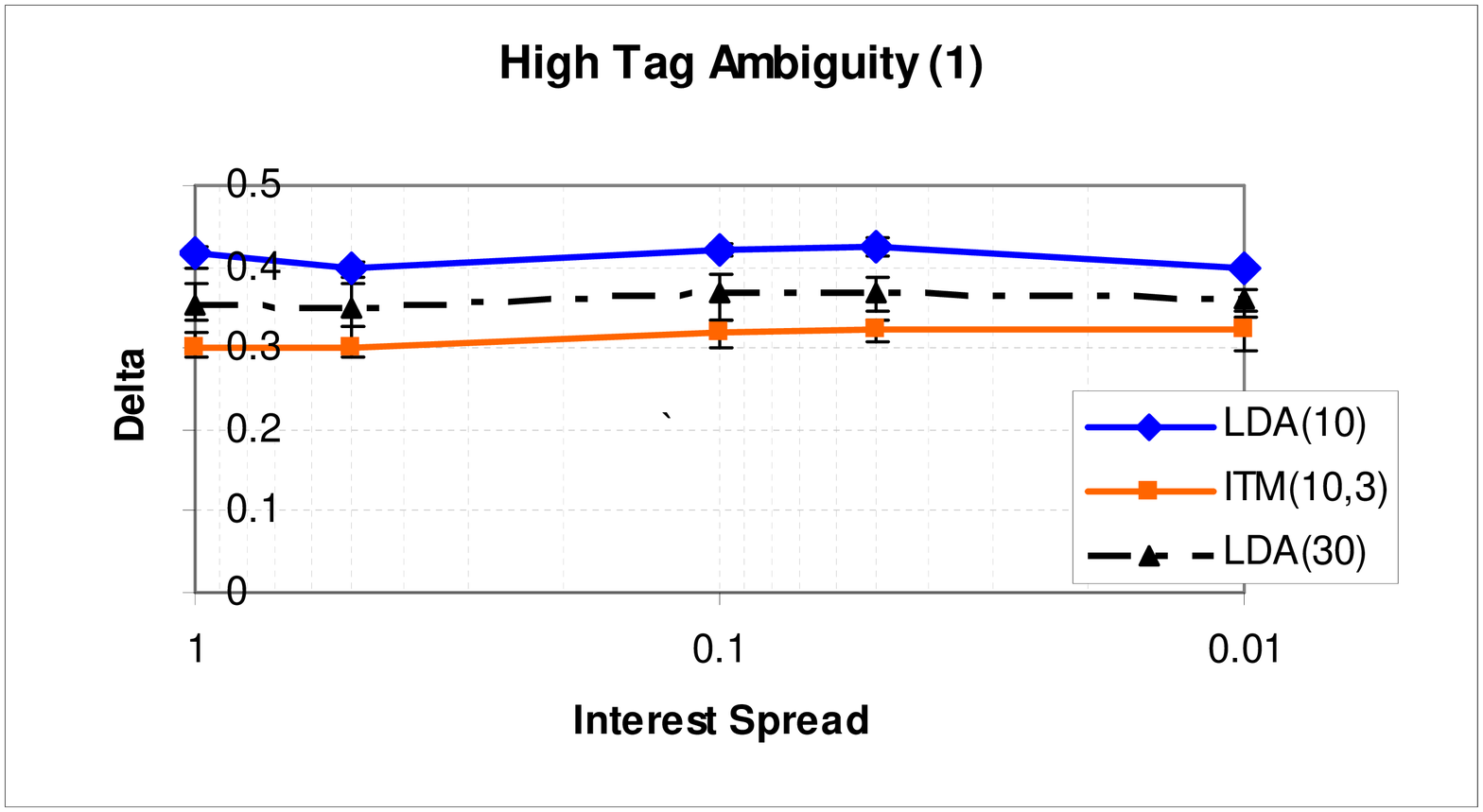} &
\includegraphics[width=2.2in]{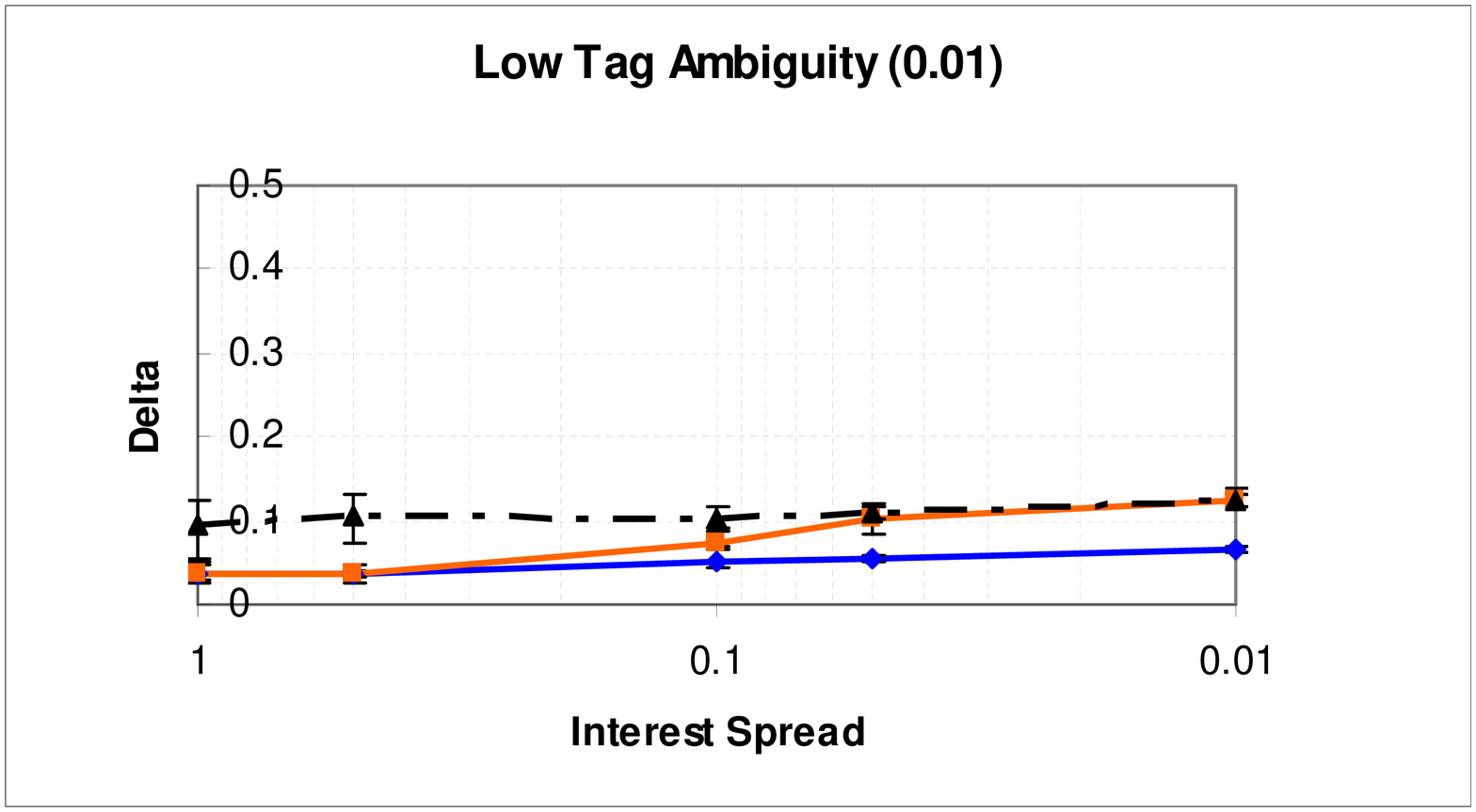} \\
(a) High Ambiguity & (b) Low Ambiguity \\
\includegraphics[width=2.2in]{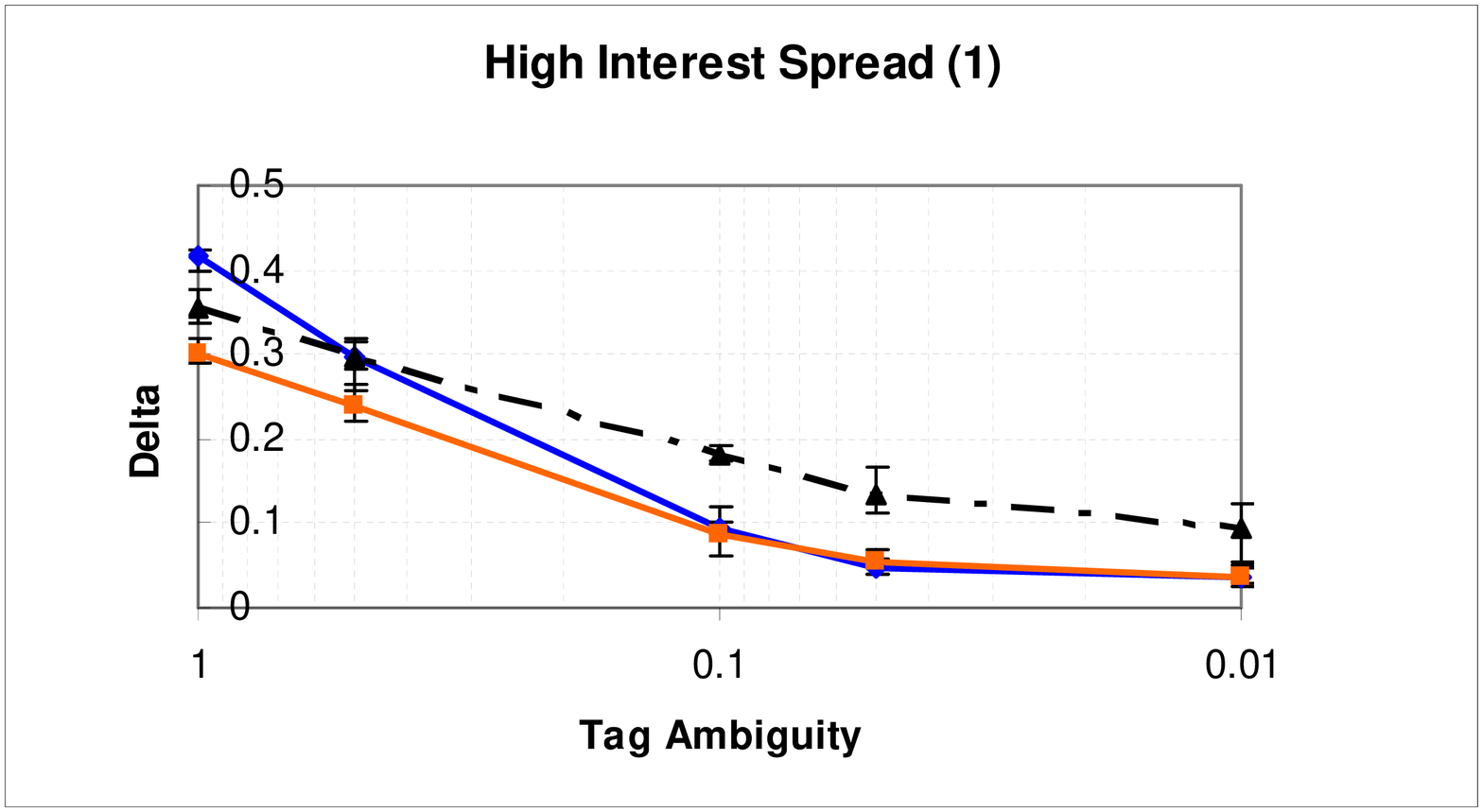} &
\includegraphics[width=2.2in]{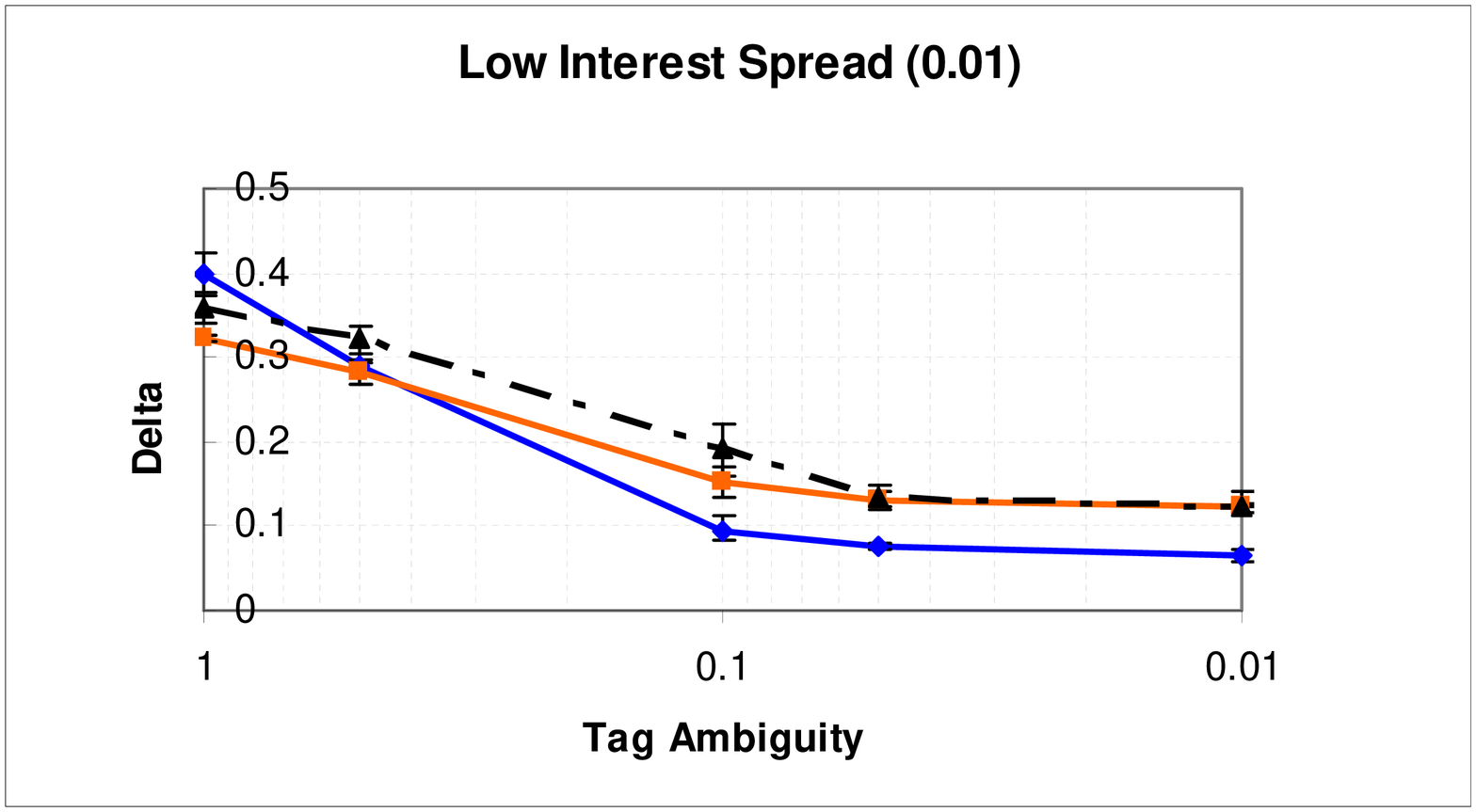} \\
(c) High Interest Spread & (d) Low Interest Spread
\end{tabular}
\end{center}
\caption {Deviations, Delta$(\Delta)$, between actual and learned topics on synthetic data sets for different regimes: (a)high tag ambiguity; (b)low tag ambiguity; (c)high interest spread; (d)low interest spread. LDA(10) and LDA(30) refers to LDA that is trained with $10$ and $30$ topics respectively; ITM(10,3) refers to ITM that is trained with $10$ topics and $3$ interests.}
\label{fig:synthetic2}
\end{figure}

We ran both LDA and ITM to learn distributions of resources over topics, $\phi$, for simulated data set generated with different values of tag ambiguity and user interest variation.  We set the number of topics to 10 for each model, and the number of interests to three for ITM. Both models were  initialized with random topic and interest assignments and then trained using 1000 iterations. For the last 100 iterations, we used topic and interest assignments in each iteration to compute $\phi$ (using ~\eqref{eq:phi} for ITM and Eq. (7) in ~\cite{GriffithsTopic04} for LDA).  The average\footnote{The reason to use the average of $\phi$ is that, in the stable state, the topic/interest assignments can still fluctuate from one iteration to another. To avoid estimate $\phi$ from an iteration that possibly has idiosyncratic topic/word assignments, one can average $\phi$ over multiple iterations \cite{TopicModel06}.} of $\phi$ in this period is then used as the distributions of resources over topics. We ran the learning algorithm five times for each data set.

% - Show experimental results
\begin{figure}
\begin{center}
\begin{tabular}{cc}
\includegraphics[width=2.4in]{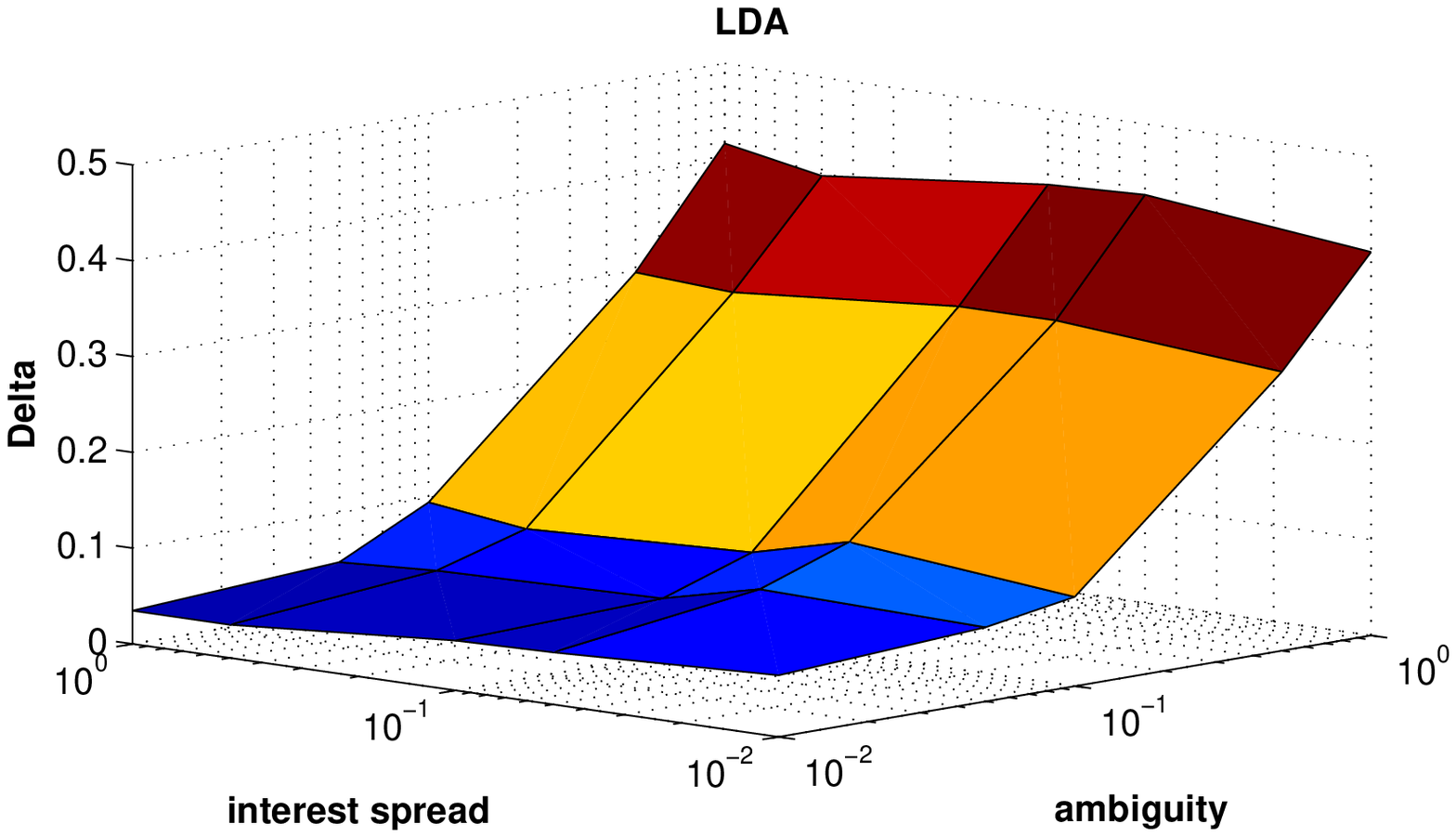}  &
\includegraphics[width=2.4in]{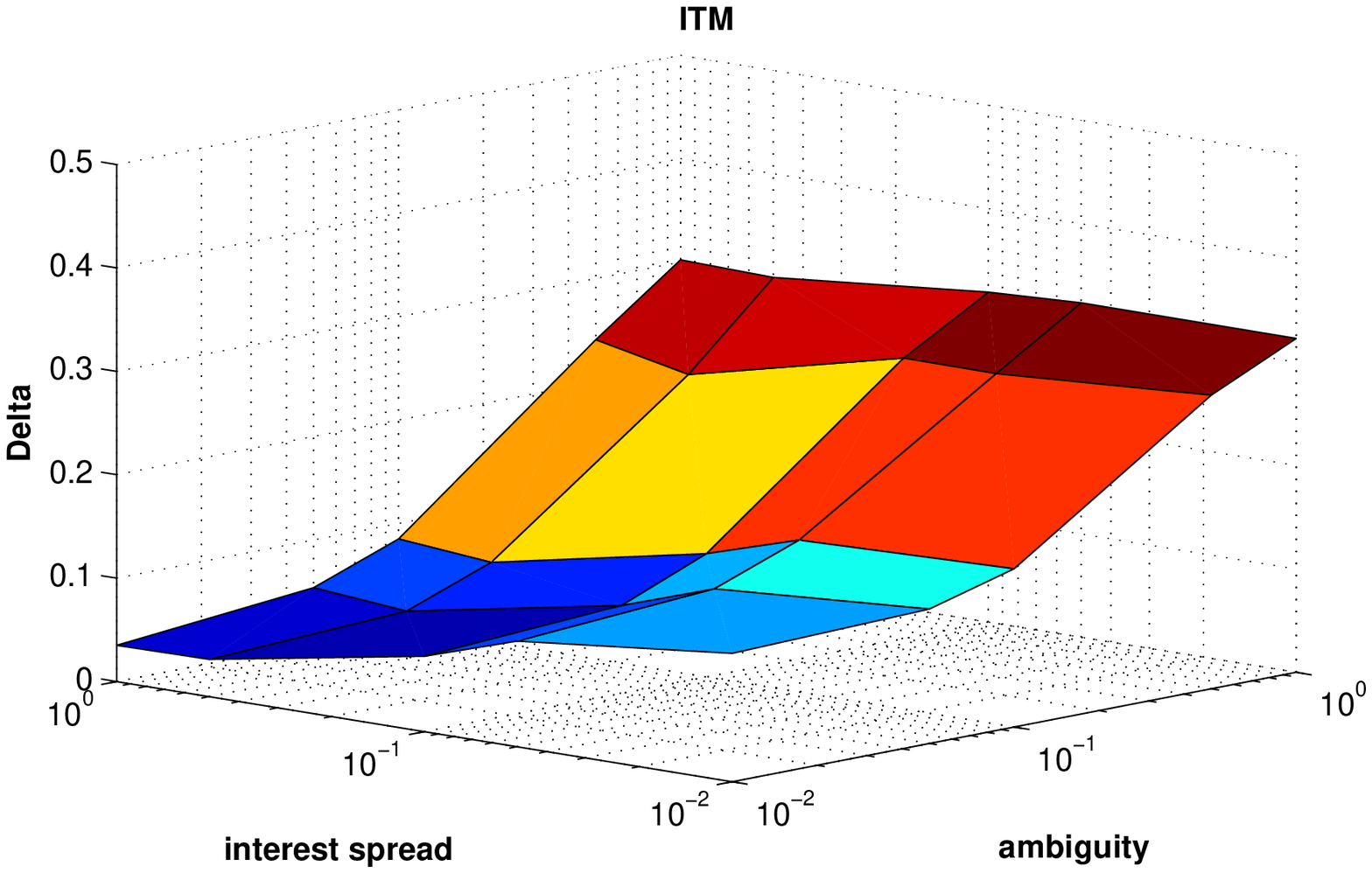} \\
(a) LDA(10) & (b) ITM(10,3)
\end{tabular}
\end{center}
\caption {This plot shows the deviation $\Delta$ between actual and learned topics on synthetic data sets, under different degrees of tag-ambiguity and user interest variation. The $\Delta$ of LDA is shown on the left (a); as that of ITM is on the right (b). The colors were automatically generated by the plotting program to improve readability.}
\label{fig:synthetic}
\end{figure}

% - Discussion on which case our model is better and which is worse.
Deviations between learned topics and actual ones of simulated data sets\comment{, generated under different conditions of tag ambiguity and user interest variations,} are shown in ~\figref{fig:synthetic2} and ~\figref{fig:synthetic}. In the case when degree of tag ambiguity is high, ITM is superior to LDA for the entire range of user interest variation, as shown in ~\figref{fig:synthetic2}(a). This is because ITM exploits user information to help disambiguate tag senses; thus, it can learn better topics, which are closer to the actual ones, than LDA. In the other regime, when tag ambiguity is low, user information does not help and can even degrade ITM performance, especially when the degree of interest variation is low, as in ~\figref{fig:synthetic2}(b). This is because low amount of user  interest variation demotes statistical strength of the learned topics.  Suppose that, for example, we have two similar resources: the first one is bookmarked by one group, the second bookmarked by another. If these two groups have very different interest profiles, ITM will tend to split the ``actual'' topic that describes those resources into two different topics --- one for each group. Hence, each of these resources will be assigned to a different learned topic, resulting in a higher  $\-{\Delta}$ for ITM.

% explain results in high ambiguity and low ambiguity regimes
In the case when user interest variation is high (\figref{fig:synthetic2}(c)), ITM is superior to LDA for  the same reason that it uses user information to disambiguate tag senses. Of course, there is no advantage to using ITM when the degree of tag ambiguity is very low, and it yields similar performance to LDA. In the last regime, when interest variation is low (\figref{fig:synthetic2}(d)), ITM is superior to LDA for high degree of tag ambiguity, even though its topics may lose some statistical strength. ITM's performance starts to degrade when tag ambiguity degree is low, for the same reason as in ~\figref{fig:synthetic2}(b). These results are summarized in 3D plots in \figref{fig:synthetic}.

We also ran LDA with 30 topics, in order to compare LDA to ITM, when both models have the same complexity\comment{ -- 10 times 3 hidden variables}. As shown in the \figref{fig:synthetic2}, with the same model complexity, ITM is preferable to LDA in all settings. In some cases, LDA with higher complexity (30 topics) is inferior to the LDA with lower complexity (10 topics). We suspect that this degradation is caused by over-specification of the model with too many topics.

%[Aug 30] computational complexity
For the computational complexity, both LDA and ITM are required to sample the hidden variables for all data points using Gibbs sampling. For LDA, only the topic variable $z$ is needed to be sampled; for ITM, the interest variable $x$ is also required. The computational cost in each sampling is proportional to a number of topics, $N_{Z}$, for $z$, and that of interest, $N_{X}$, for $x$. Let define $\kappa$ as a constant. We also define a number of all datapoints (tuples) as $N_{K}$. Hence, a computational cost for LDA, in each iteration can be approximated as $N_{K} \times (\kappa \times N_{Z})$. The computational cost of ITM in each iteration can be approximated as $N_{K} \times (\kappa \times (N_{Z} + N_{X}))$.

In summary, ITM is not superior to LDA in learning topics associated with resources in every case. However, we showed that ITM is preferable to LDA in scenarios characterized by a high degree of tag ambiguity and some user interest variation, which is the case in the social annotations domain.

%AP120509
\subsection{Real-World Data}
\label{sec:realitm}
In this section we validate the proposed model on real-world data obtained from the social bookmarking site \emph{Delicious}. The hypothesis we make for evaluating the proposed approach is that the model that takes users into account can infer higher quality (more accurate) topics $\phi$ than those inferred by the model that ignores user information.

The ``standard'' measure\footnote{In fact, topic model's evaluation is still currently in controversy according to a personal communication at \emph{http://nlpers.blogspot.com/2008/06/evaluating-topic-models.html} by Hal Daum\'{e}.} used for evaluating topic models is the perplexity score \cite{BleiLDA03,TopicModelSmyth2004}. Specifically, it measures generalization performance on how a certain model can predict unseen observations. In document topic modeling, a portion of words in each document are set aside as testing data; while the rest are used as training data. Then the perplexity score is computed from a conditional probability of the testing given training data. This evaluation is infeasible in the social annotation domain, where each bookmark contains relatively few tags, compared to document's words.

Instead of using perplexity, we propose to directly measure the quality of the learned topics on a simplified resource discovery task. The task is defined as follows: ``given a \emph{seed} resource, find other most similar resources''~\cite{AmbiteISWC09}. Each resource is represented as a distribution over learned topics, $\phi$, which is computed using \eqref{eq:phi}. Topics learned by the better approach will have more discriminative power for categorizing resources. When using such distribution to rank resources by similarity to the seed, we would expect the more similar resources to be ranked higher than less similar resources. Note that similarity between a pair of resources $A$ and $B$ is computed using Jensen-Shannon divergence (JSD)~\cite{JSDiverge} on their topic distributions, $\phi_{A}$ and $\phi_{B}$.

To evaluate the approach, we collected data  for five seeds:
\emph{flytecomm},\footnote{http://www.flytecomm.com/cgi-bin/trackflight/}
\emph{geocoder},\footnote{http://geocoder.us}
\emph{wunderground},\footnote{http://www.wunderground.com/}
\emph{whitepages},\footnote{http://www.whitepages.com/} and
\emph{online-reservationz}.\footnote{http://www.online-reservationz.com/}
The \emph{flytecomm} allows users to track flights given the airline and flight number or departure and arrival airports; \emph{geocoder} returns geographic coordinates of a given address; \emph{wunderground} gives weather information for a particular location (given by zipcode, city and state, or airport); \emph{whitepages} returns person's phone numbers and \emph{online-reservationz} lists hotels available in some city on some dates.
\comment{We exploits the tagging activity of \emph{Delicious} users to retrieve the corpus associated with each seed.} We crawl \emph{Delicious} to gather resources possibly relating to each seed. The crawling strategy is as follows: for each seed
\begin{itemize}
\item retrieve the 20 most popular tags associated with this resource.

\item For each of the tags, retrieve other resources that have been annotated with the tag.

\item For each resource, collect all bookmarks (resource-user-tag triples).

\end{itemize}
We wrote a special-purpose page scraper to extract this information from \emph{Delicious}. In principle, we could continue to expand the collection of resources by gathering tags and retrieving more resources tagged with those keywords, but in practice, even after a small traversal, we already obtain millions of triples.
In each corpus, each resource has at least one tag in common with the seed. Statistics on these data sets are presented in \tabref{tbl:deliciousstats}.

% table on data set statistics
\begin{table}
\begin{tabular}{|l|l|l|l|l|}
\hline
Seed & \# Resources & \# Users & \# Tags & \#Tripples \\
\hline
Flytecomm & 3,562 & 34,594 & 14,297 & 2,284,308 \\
\hline
Geocoder & 5,572 & 46,764 & 16,887 & 3,775,832 \\
\hline
Wunderground & 7,176 & 45,852 & 77,056 & 6,327,211 \\
\hline
Whitepages & 6,455 & 12,357 & 64,591 & 2,843,427 \\
\hline
Online-Resevationz & 764 & 41,003 & 9,194 & 162,763 \\
\hline
\end{tabular}
\caption{The table presents statistics for five data sets for evaluating models' performance. Note that a triple is a resource, user, and tag co-occurrence.}
\label{tbl:deliciousstats}
\end{table}
% say something about evaluation

For each corpus, LDA is trained with 80 topics, while the number of topics and interests for ITM is set to $80$ and $40$ respectively. The topic and interest assignments are randomly initialized, and then both models are trained with the $500$ iterations.\footnote{We discovered that the model converging very quickly. In particular, the model appear to reach the stable state within $300$ iterations in all data sets} For the last 100 iterations, we use the topic and interest assignments, in each iteration, to compute the distributions of resources over topics, $\phi$. The average of $\phi$ in this period is then used as the distributions of resources over topics.

Next, the learned distributions of resources over topics, $\phi$, are used to compute the similarity of resources in each corpus to the seed. The performance of each model is evaluated by manually checking the $100$ most similar resources produced by the model. \comment{The ground truth is as follows: the}A resource is judged to be similar if it provides an input form that takes semantically the same inputs as the seed and returns semantically the same data. Hence, \emph{flightaware}\footnote{http://flightaware.com/live/} is judged similar to \emph{flytecomm} because both take flight information and return flight status.

% places results here

\begin{figure}
\begin{center}
\begin{tabular}{cc}
\includegraphics[width=2.2in]{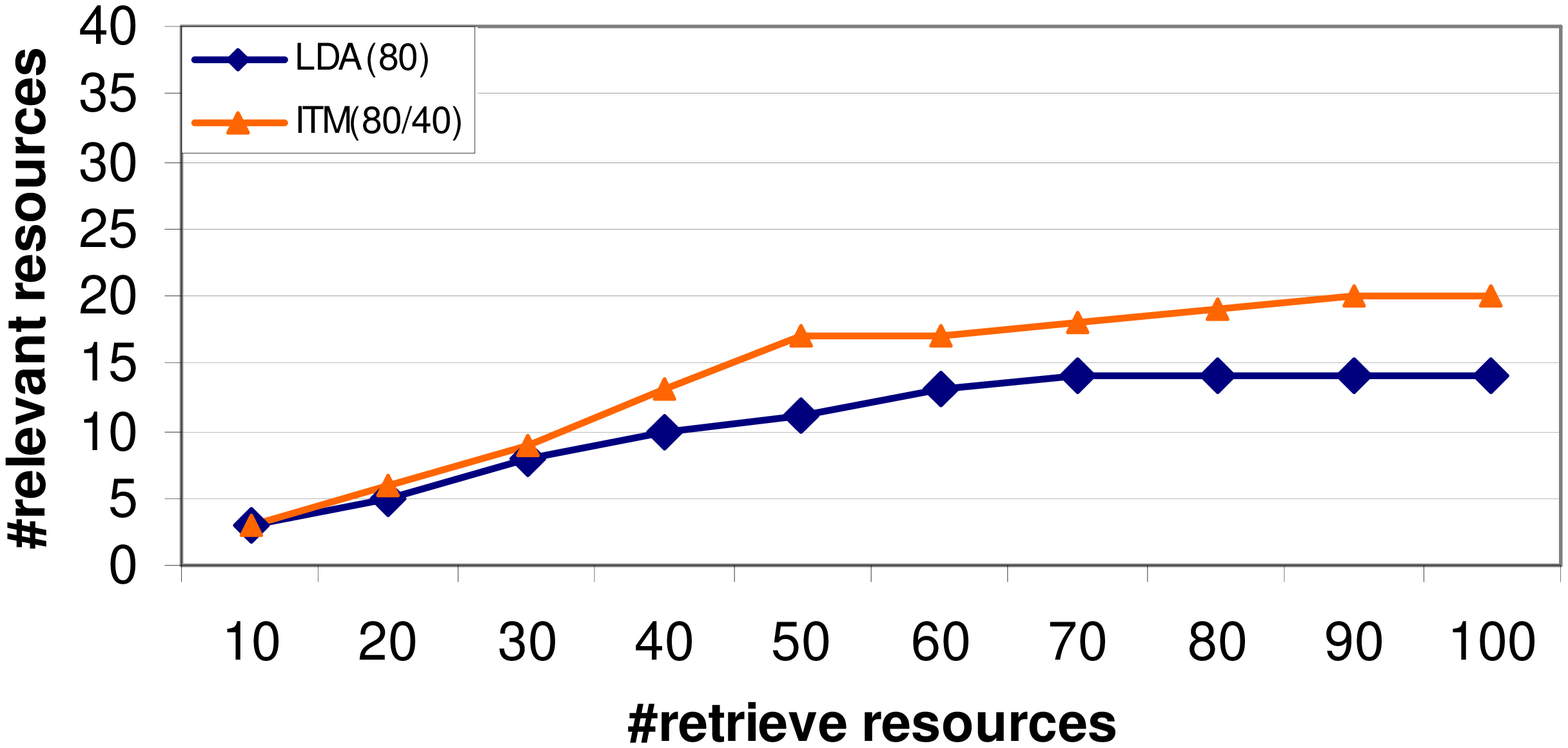} &
\includegraphics[width=2.2in]{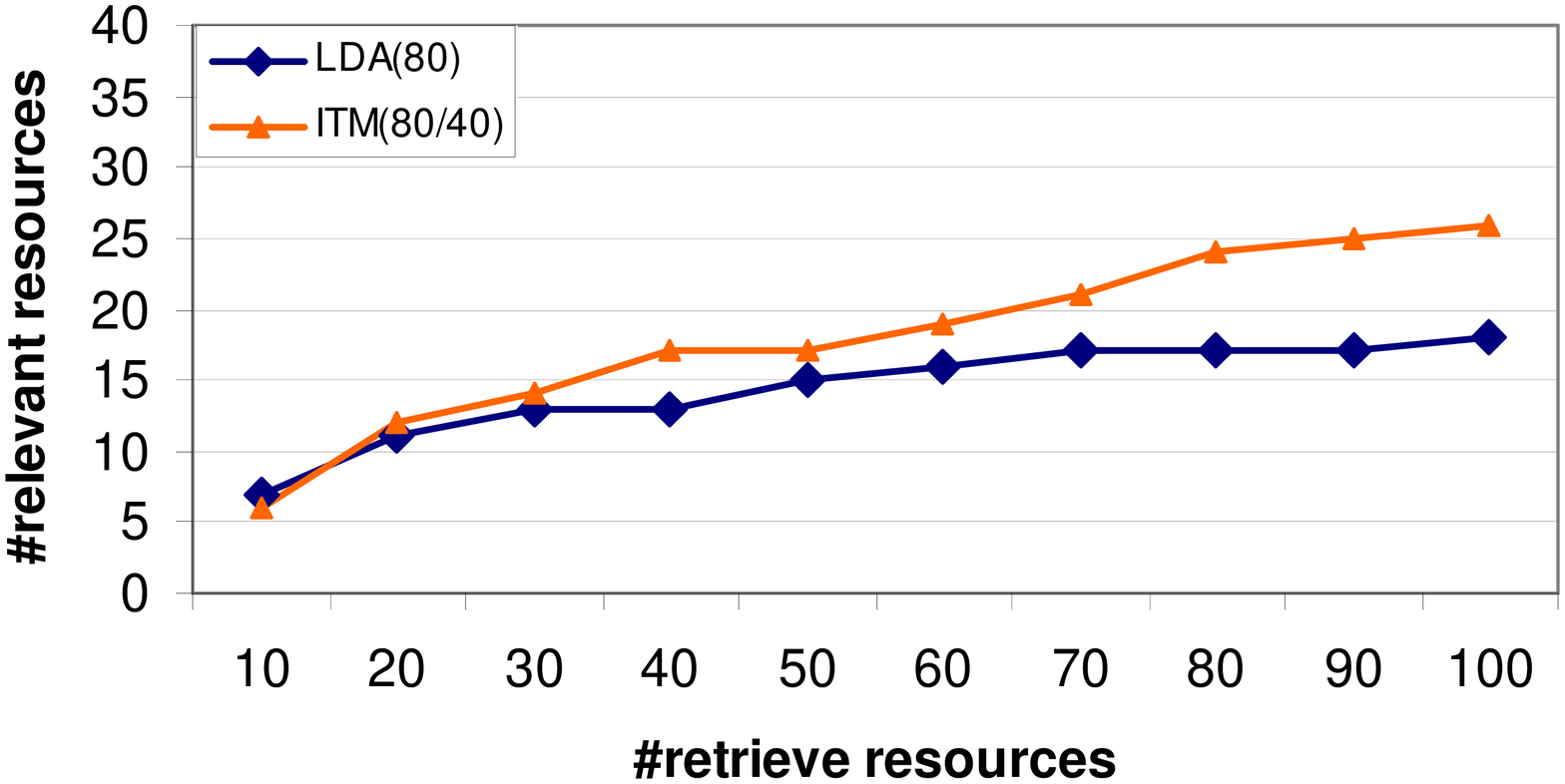} \\
(a) Flytecomm & (b) Geocoder \\
\includegraphics[width=2.2in]{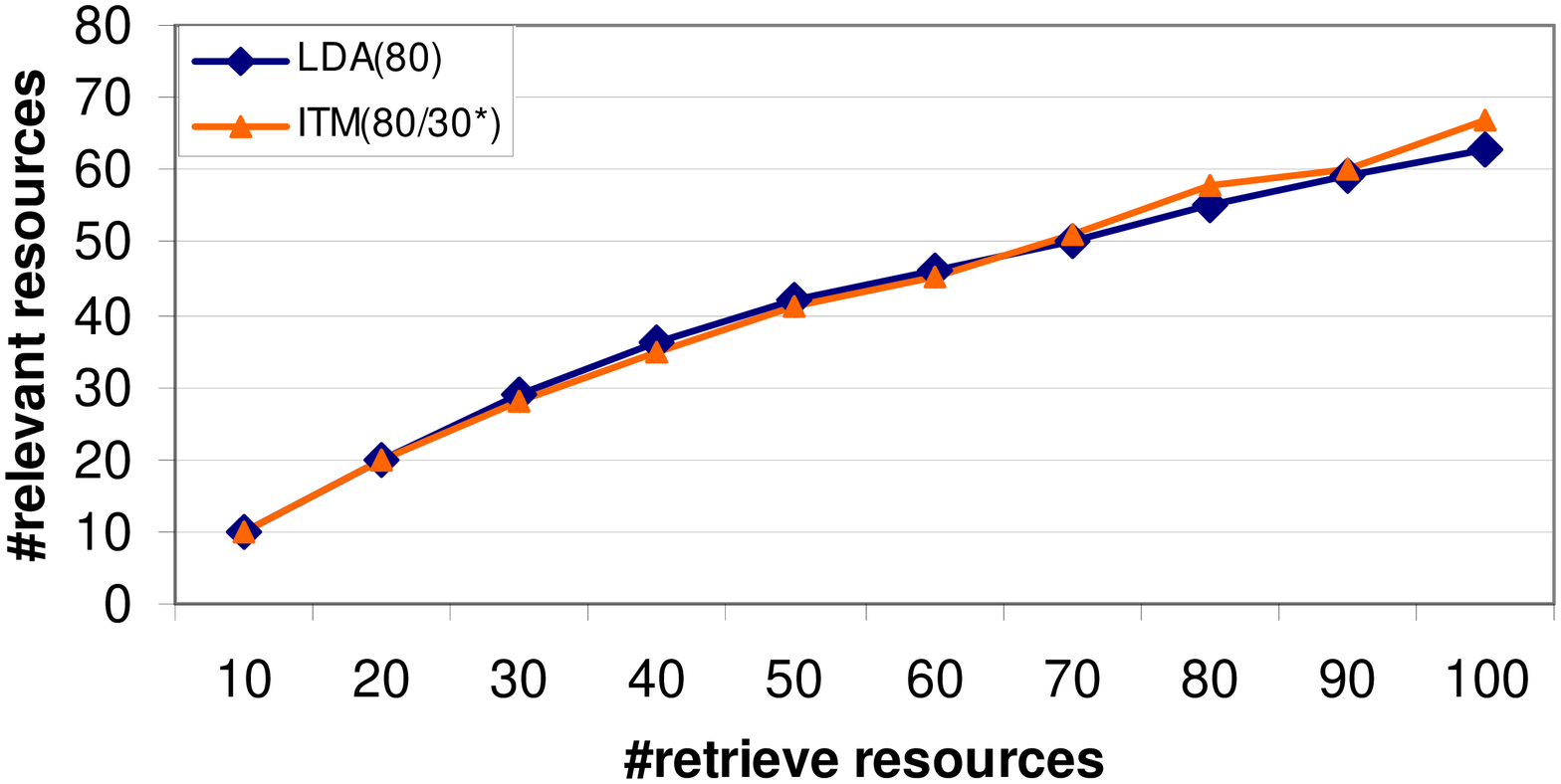} &
\includegraphics[width=2.2in]{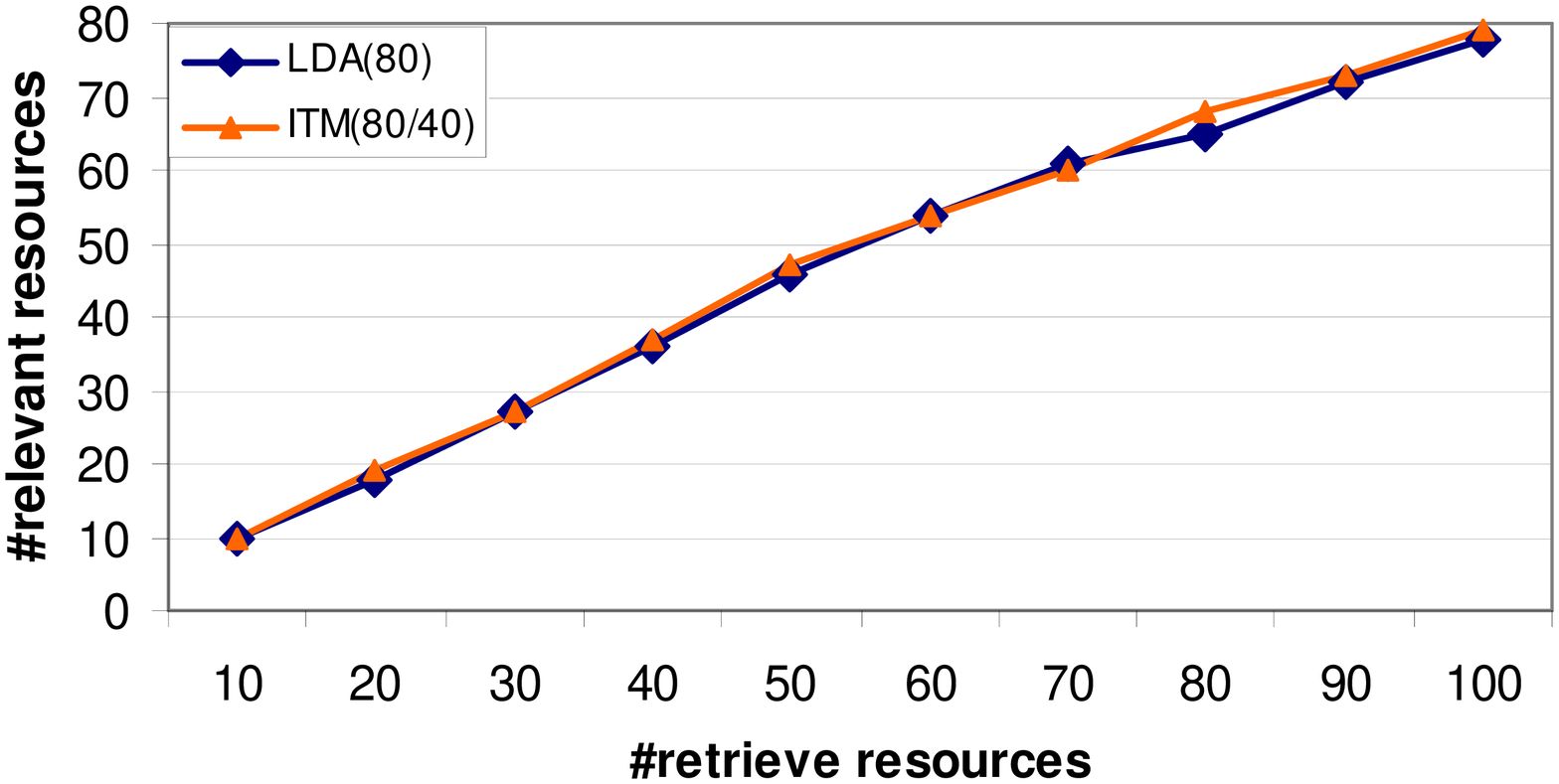} \\
(c) Wunderground & (d) Whitepages\\
\includegraphics[width=2.2in]{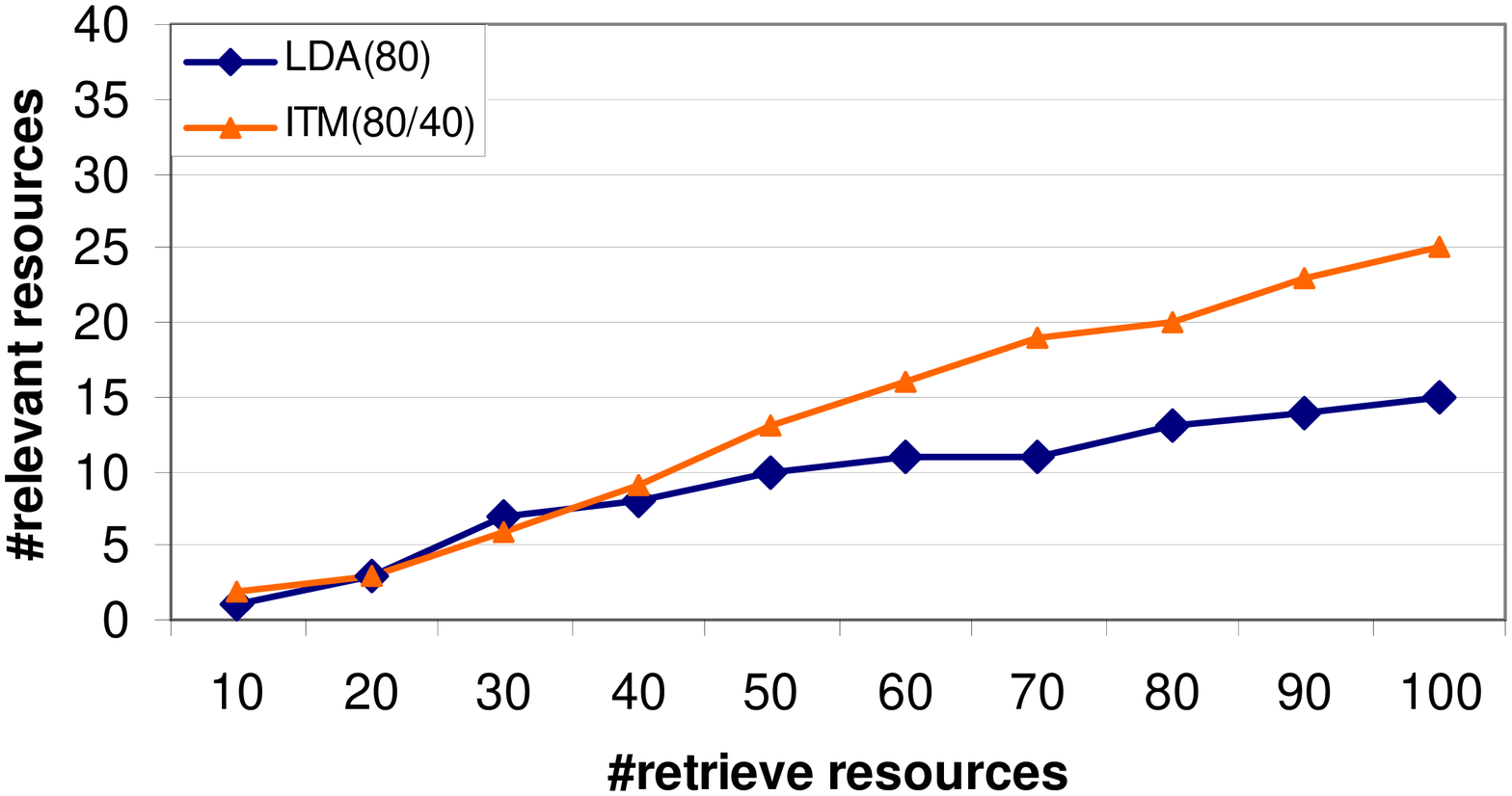} &
\\
(e) Online-reservationz & \\

\end{tabular}
\end{center}
\caption {Performance of different models on the five data sets. X-axis represents the number of retrieved resources; y-axis represents the number of relevant resources (that have the same function as the seed). LDA(80) refers to LDA that is trained with $80$ topics. ITM(80/40) refers to ITM that is trained with $80$ topics and $40$ interests. In \emph{wunderground} case, we can only run ITM with 30 interests due to the memory limits.}
\label{fig:rankresults1}
\end{figure}

\figref{fig:rankresults1} shows the number of relevant resources identified within the top $x$ resources returned by LDA and ITM. From the results, we can see that ITM is superior to LDA in three data sets: \emph{flytecomm}, \emph{geocoder} and \emph{online-reservationz}. However, its performance for \emph{wunderground} and \emph{whitepages} is about the same as that of LDA.  Although we have no empirical proof, we hypothesize that weather and directory services are of interest to all users, and are therefore bookmarked by a large variety of users, unlike users interested in tracking flights or booking hotels online\comment{, nevertheless, are a special subgroup of users}. As a result, ITM cannot exploit individual user differences to learn more accurate topics $\phi$ in the \emph{wunderground} and \emph{whitepages} cases.

To illustrate utility of ITM, we select examples of topics and interests of the model induced from the \emph{flytecomm} corpus. For purposes of visualization, we first list in descending order the top tags that are highly associated with each \emph{topic}, which are obtained from $\theta_{z}$ (aggregated over all interests in the topic $z$). For each topic, we then enumerate some \emph{interests}, and present a list of top tags for each interest, obtained from $\theta_{x,z}$. We manually label topics and interests (in \textit{italics}) according to the meaning of its dominant tags.

\begin{quotation}

\noindent \textit{Travel \& Flights topic}: \texttt{travel, Travel, flights, airfare, air\-line, flight, air\-lines, guide, aviation, hotels, deals, re\-ference, air\-plane}
\begin{itemize}
\item \textit{Flight Tracking interest}: \texttt{travel, flight, airline, airplane, tracking, guide, flights, hotel, aviation, tool, \\packing, plane}
\item \textit{Deal \& Booking interest}: \texttt{vacation, service, travelling, hotels, search, deals, europe, portal, tourism, price, compare, old}
\item \textit{Guide interest}: \texttt{travel, cool, useful, reference, world, advice, holiday, international, vacation, guide,} \\ \texttt{information, resource} \\
\end{itemize}

\noindent \textit{Video \& p2p topic}: \texttt{video, download, bittorrent, p2p, youtube, media, torrent, torrents, movies, videos, Video,  \\downloads, dvd, free, movie}
\begin{itemize}
\item \textit{p2p Video interest}: \texttt{video, download, bittorrent, youtube, torrents,	p2p, torrent, videos, movies, dvd, media, googlevideo, downloads, pvr}
\item \textit{Media \& Creation interest}: \texttt{video, media, movies, multimedia, videos, film, editing, vlog, remix, sharing, rip, \\ipod, television, videoblog}
\item \textit{Free Video interest}: \texttt{video, free, useful, videos, cool, \\downloads, hack, media, utilities, tool, hacks, flash, audio, podcast}\\

\end{itemize}

\noindent \textit{Reference topic}: \texttt{reference, database, cheatsheet, Reference, resources, documentation, list, links, sql, lists, \\resource, useful, mysql}
\begin{itemize}
\item \textit{Databases interest}: \texttt{reference, database, documentation, sql, info, databases, faq, technical, reviews, tech, oracle, manuals}
\item \textit{Tips \& Productivity interest}: \texttt{reference, useful, resources, \\information, tips, howto, geek, guide, info, produc\-ti\-vity, daily, computers}
\item \textit{Manual \& Reference interest}: \texttt{resource, list, guide, resources, collection, help, directory, manual, index, portal, archive, bookmark}

\end{itemize}

\end{quotation}

%AP120 describe the list of topics and interests above
The three interests in the ``Travel \& Flights'' topic have obviously different themes. The dominant one is more about tracking status of a flight; while the less dominant ones are about searching for travel deals and traveling guides respectively. This implies that there are subsets of users who have different perspectives (or what we call interests) towards the same topic. Similarly, different interests also appear in the following topics, ``Video \& p2p'' and ``Reference.''

\begin{figure}
\begin{center}
\begin{tabular}{c}

\includegraphics[width=4.0in]{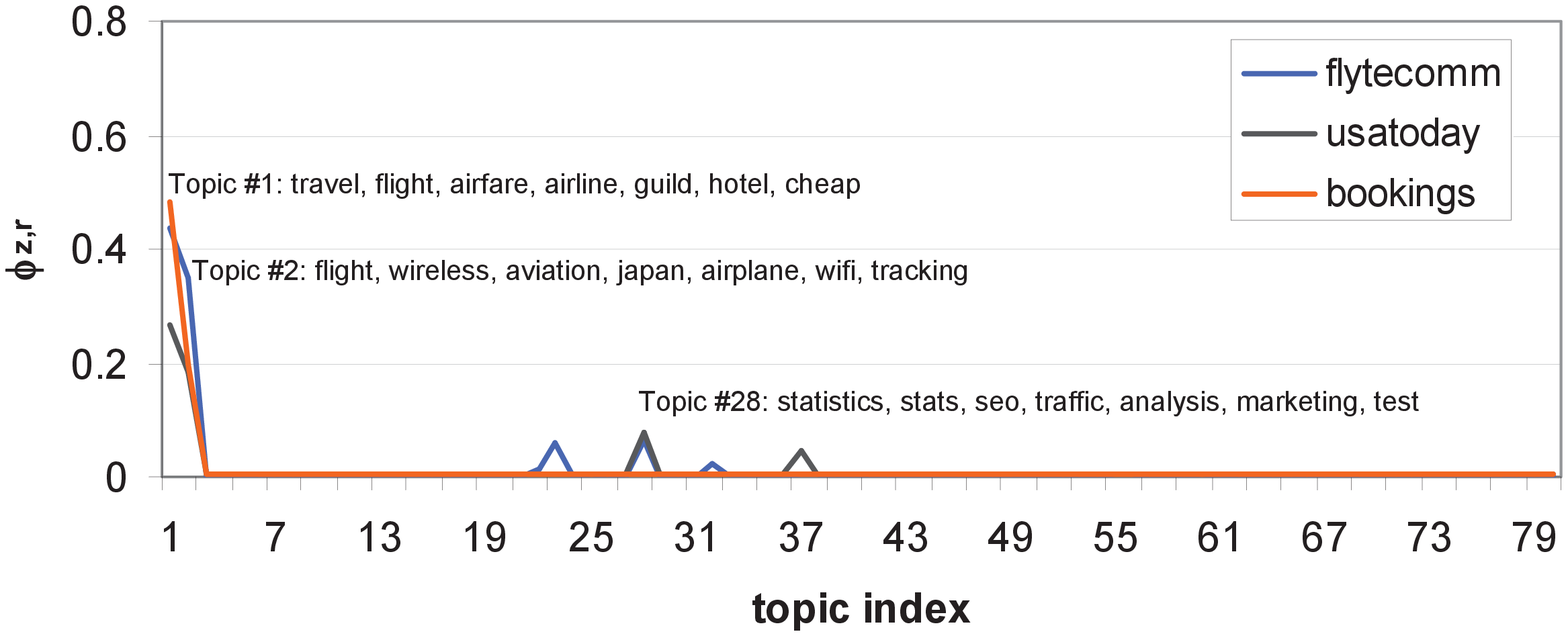} \\%{charts/topicdist_lda.eps}  \\
(a) Tag distributions of three resources learned by LDA \\
\includegraphics[width=4.0in]{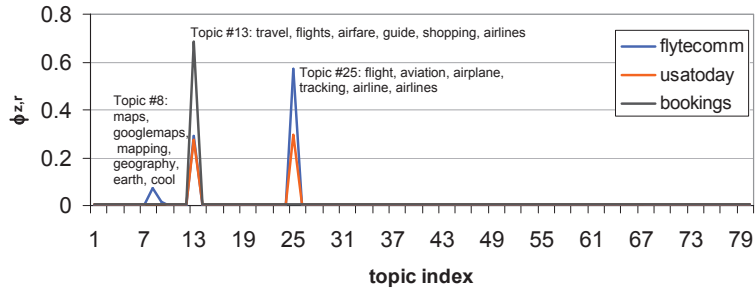}\\%{charts/topicdist_itm.eps} \\
(b) Tag distributions of three resources learned  by ITM 

\end{tabular}
\end{center}
\caption{Topic distributions of three resources: \emph{flytecomm}, \emph{usatoday},\comment{\footnote{http://www.usatoday.com/travel/flights/delays/tracker-index.htm
}} \emph{bookings}\comment{\footnote{http://www.bookings.org/}} learned by (a) LDA and (b) ITM. $\phi_{z,r}$ in y-axis indicates a weight of the topic $z$ in the resource $r$ -- the degree to which $r$ is about the topic $z$.}
\label{fig:tagdist}
\end{figure}

\figref{fig:tagdist} presents examples of topic distributions for three resources learned by LDA and ITM: the seed \emph{flytecomm}, \emph{usatoday},\footnote{http://www.usatoday.com/travel/flights/delays/tracker-index.htm
} and \emph{bookings}.\footnote{http://www.bookings.org/} Although all are about travel, the first two resources have specific flight tracking functionality; while the last one is about hotel \& trip booking. In distribution of resources over the topics learned by LDA, shown in \figref{fig:tagdist} (a), all resources have high weights on topics \#1 and \#2, which are about traveling deals and general aviation. In the case of topics learned by ITM, shown in \figref{fig:tagdist} (b), \emph{flytecomm} and \emph{usatoday} have their high weight on topic \#25, which is about tracking flights, while \emph{bookings} does not. Consequently, ITM will be more helpful than LDA in identifying flight tracking resources. This demonstrates the advantage of ITM in exploiting individual differences to learn more accurate topics.

\section{Infinite Interest Topic Model} % with Hierarchical Dirichlet Process}
\label{sec:infiniteitm}

In \secref{sec:finiteitm}, we assumed that parameters, such as, $N_{Z}$ and $N_{X}$ (number of topics and interests respectively), were fixed and known \emph{a priori}. The choice of values for these parameters can conceivably affect the model performance. The traditional way to determine these numbers is to learn the model several times with different values of parameters, and then select those that yield the best performance \cite{GriffithsTopic04}.

In this work, we choose another solution by extending our finite model to have ``countably'' infinite numbers of topics and interests. By ``countably'' infinite number of components, we mean that such numbers are flexible and can vary according to the number of observations. Intuitively, there is a higher chance that more topics and interests will be found in a data set that has more resources and users. Such unbounded number of components can be dealt with within a Bayesian framework, as mentioned in the previous works \cite{NealDPM00,InfiniteGM00,Teh04hierarchicaldirichlet}. This approach helps bypass the problem of selecting values for these parameters.

Following \cite{NealDPM00}, we set both $N_{Z}$ and $N_{X}$ to approach $\infty$. This will give the model the ability to select not only previously used topic/interest components but also to instantiate ``unused'' components when required. However, the model that we derived in the previous section cannot be extended directly under this framework due to the use of symmetric Dirichlet priors. As pointed out by \cite{Teh04hierarchicaldirichlet}, when the number of components grows,
% KL - the 'size' or the 'number' above?
% [Aug 27] I changed it to number
using the symmetric Dirichlet prior results in a very low --- even zero probability --- chance that a mixture component is shared across groups of data. That is, in our context, there is a higher chance that a certain topic is only used within one resource rather than utilized by many of them. Considering \eqref{eq:gibbsZ}, if we set $N_{Z}$ to approach $\infty$, we can obtain posterior probability of $z$ as follows

\begin{eqnarray}
p(z_{i}=z_{used}|\textbf{z}_{-i},\textbf{x},\textbf{t}) = \frac{N_{r_{i},z_{-i}}}{N_{r_{i}}+\alpha-1}\cdot\frac{N_{z_{-i},x_{i},t_{i}}+\eta/N_{T}}{N_{z_{-i},x_{i}}+\eta}
\label{eq:infgibbsZused}
\end{eqnarray}

\begin{eqnarray}
p(z_{i}=z_{new}|\textbf{z}_{-i},\textbf{x},\textbf{t}) = \frac{\alpha}{N_{r_{i}}+\alpha-1}\cdot\frac{1}{N_{T}}
\label{eq:infgibbsZnew}
\end{eqnarray}

From \eqref{eq:infgibbsZused}, we can perceive that the model only favors topic components that are only used within the resource $r_{i}$.
Meanwhile, for other components that are not used by that resource, $N_{z_{-i},x_{i},t_{i}}$ would equal zero and thus result in zero probability in choosing them. Consequently, the model only chooses topic components for a resource either from components that are currently used by that resource, or it instantiates a new component for that resource with probabilities according to \eqref{eq:infgibbsZused} and \eqref{eq:infgibbsZnew} respectively. As more new components are instantiated, each resource tends to own its components exclusively. From the previous section, we can also perceive that each resource profile is generated independently (using symmetric Dirichlet prior) --- there is no mechanism to link the used components across different resources\footnote{This behavior can be easily observed in multiple samples, each drawn independently from a Dirichlet distribution $Dirichlet(\alpha_{1},...,\alpha_{k})$ . If $\alpha_{i}$ is ``small'' and $k$ is ``large'', there is a higher chance that samples obtained from this Dirichlet distribution will have no overlapped component i.e., for any pair of samples, there is no case when the same components have their value greater than 0 at the same time. Lack of this component overlap across samples will be obvious when $k \rightarrow \infty$. This is the problem that can be found in the model with infinite limit on $N_{Z}$ and $N_{X}$.}.
As mentioned in \cite{Teh04hierarchicaldirichlet}, this is an undesired characteristic, because, in our context, we would expect ``similar'' resources to be described by the same set of ``similar'' topics. \commentout{Note that this problem does not occur in the case that we fix a number of components since such contraint forces the model always choosing one of existing in-use components.}

\begin{figure}
\begin{center}
\includegraphics[width=3.5in]{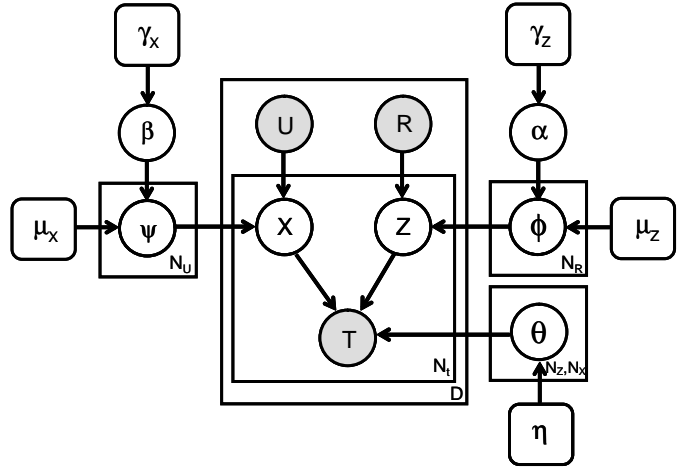}
\end{center}
\caption {Graphical representation on the Interest Topic model with hierarchical Dirichlet process (HDPITM).
 } \label{fig:hdpitm}
\end{figure}

One possible way to handle this problem is to use Hierarchical Dirichlet Process (HDP) \cite{Teh04hierarchicaldirichlet} as the prior instead of the symmetric Dirichlet prior. The idea of HDP is to link components at group-specific level together by introducing global components across all groups. Each group is only allowed to use some (or all) of these global components and thus, some of them are expected to be shared across several groups. We adapt this idea by considering all tags of resource $r$ to belong to the resource group $r$.
 Similarly, all tags of user $u$ belong to the user group $u$. Each of the resource groups is assigned to some topic components selected from the global topic component pool. Similarly, each of the user groups is assigned to some  interest components selected from the global interest component pool. This extension is depicted in \figref{fig:hdpitm}.  Suppose that a number of all possible topic components is $N_{Z}$ (which will be set to approach $\infty$ later on) and that for interest components is $N_{X}$, we can describe such extension as a stochastic process as follows.

At the global level, the weight distribution of components is sampled according to
\begin{description}
\item $(\beta_{1},...,\beta_{N_{X}}) \sim Dirichlet(\gamma_{x}/N_{X},...,\gamma_{x}/N_{X})$ (generating global interest component weight)
\item $(\alpha_{1},...,\alpha_{N_{Z}}) \sim Dirichlet(\gamma_{z}/N_{Z},...,\gamma_{z}/N_{Z})$ (generating global topic component weight)
\end{description}
\noindent where  $\gamma_{x}$ and $\gamma_{z}$ are concentration parameter, which controls diversity of interests and topics at global level.

At the group specific level,
\begin{description}
\item $\psi_{u} \sim Dirichlet(\mu_{x}\cdot\beta_{1},...,\mu_{x}\cdot\beta_{N_{X}})$ (generating user $u$ interest's profile)
\item $\phi_{r} \sim Dirichlet(\mu_{z}\cdot\alpha_{1},...,\mu_{z}\cdot\alpha_{N_{Z}})$ (generating resource $r$ topic's profile)
\end{description}
\noindent where $\mu_{x}$ and $\mu_{z}$ are concentration parameter, which controls diversity of interests and topics at group specific level. The remaining steps involving generation of tags for each bookmark are the same as in the previous process.

Suppose that there is an infinite number of all possible topics, $N_{Z} \rightarrow \infty$, and a portion of them are currently used in some resources. By following \cite{Teh04hierarchicaldirichlet}, we can rewrite the global weight distribution of topic components, \textbf{$\alpha$}, as  $(\alpha_{1},..\alpha_{k_{z}},\alpha_{u})$, where $k_{z}$ is the number of currently used topics components and $\alpha_{u} = \sum_{k=k_{z}+1}^{N_{Z}}{\alpha_{k}}$ --  all of unused topic components. Similarly, we can write $(\alpha_{1},...,\alpha_{k_{z}},\alpha_{u}) \sim Dirichlet(\gamma_{z/{N_{Z}}},...,\gamma_{z/{N_{Z}}},\gamma_{z_{u}})$, where $\gamma_{z/{N_{Z}}} = \gamma_{z}/N_{z}$ and $\gamma_{z_{u}} = \frac{(N_{Z}-k_{z})\cdot\gamma_{z}}{N_{Z}}$. The same treatment is also applied to that of interest components.

Now we can generalize \eqref{eq:gibbsZ} and \eqref{eq:gibbsX} for sampling posterior probabilities of topic $z$ and interest $x$ with HDP priors as follows.

For sampling topic component assignment for datapoint $i$,
\begin{eqnarray}
p(z_{i}=k|\textbf{z}_{-i},\textbf{x},\textbf{t}) = \frac{N_{r_{i},z_{-i}}+\mu_{z}\alpha_{k}}{N_{r_{i}}+\mu_{z}-1}\cdot\frac{N_{z_{-i},x_{i},t_{i}}+\eta/N_{T}}{N_{z_{-i},x_{i}}+\eta}
\label{eq:gibbsZnpUsed}
\end{eqnarray}

\begin{eqnarray}
p(z_{i}=k_{new}|\textbf{z}_{-i},\textbf{x},\textbf{t}) = \frac{\mu_{z}\alpha_{u}}{N_{r_{i}}+\mu_{z}-1}\cdot\frac{1}{N_{T}}
\label{eq:gibbsZnpNew}
\end{eqnarray}

For sampling interest component assignment for datapoint $i$,
\begin{eqnarray}
p(x_{i}=j|\textbf{x}_{-i},\textbf{z},\textbf{t}) = \frac{N_{u_{i},x_{-i}}+\mu_{x}\beta_{j}}{N_{u_{i}}+\mu_{x}-1}\cdot\frac{N_{x_{-i},z_{i},t_{i}}+\eta/N_{T}}{N_{x_{-i},z_{i}}+\eta}
\label{eq:gibbsXnpUsed}
\end{eqnarray}

\begin{eqnarray}
p(x_{i}=j_{new}|\textbf{x}_{-i},\textbf{z},\textbf{t}) = \frac{\mu_{x}\beta_{u}/N_{X}}{N_{u_{i}}+\beta-1}\cdot\frac{1}{N_{T}}\,,
\label{eq:gibbsXnpNew}
\end{eqnarray}
\noindent where $k$ and $j$ are an index for topic and interest component respectively. From these equations, we allow the model to instantiate a new component from the pool of unused components. Considering the case when a new topic component is instantiated and, for simplicity, we set this new component to be the last used component, indexed with $k_{z}^{\prime}$. We need to obtain weight $\alpha_{k_{z}^{\prime}}$ for this new component and also update the weight of all unused components, $\alpha_{u^{\prime}}$. From the unused component pool, we know that one of its unused components will be chosen as a newly used component, $k_{z}^{\prime}$, with probability distribution $(\alpha_{k_{z}+1}/\alpha_{u},..,\alpha_{N_{Z}}/\alpha_{u})$ which can be sampled from $Dirichlet(\gamma_{z}/N_{Z},...,\gamma_{z}/N_{Z})$. \commentout{(according to the component's generative process at global level).} Suppose the component $k_{z}^{\prime}$ will be chosen from one of these components and we collapse the remaining unused components. It will be chosen with the probability $\alpha_{k_{z}^{\prime}}/\alpha_{u}$, which can be sampled from $Beta(\gamma_{z}/N_{Z},\gamma_{z_{u}}/N_{Z}-\gamma_{z}/N_{Z})$, where $Beta(.)$ is a Beta distribution.

% KL [Sep 4] Define Beta (it's Beta distribution -- the dirichlet distribution with only 2 instead of n parameters)

Now, suppose $k_{z}^{\prime}$ is chosen. The probability of choosing this component is updated to $\alpha_{k_{z}^{\prime}}/\alpha_{u} \sim Beta(\gamma_{z}/N_{Z}+1,\gamma_{z_{u}}/N_{Z}-\gamma_{z}/N_{Z})$ \commentout{(a basic property of Dirichlet and Beta distribution, I use this from \cite{Teh04hierarchicaldirichlet} and Aaron D' Souza notes: http://www-clmc.usc.edu/~cs599_ct/dirichlet_processes.pdf)}. When $N_{Z} \rightarrow \infty$, this reduces to $\alpha_{k_{z}^{\prime}}/\alpha_{u} \sim Beta(1,\gamma_{z_{u}}/N_{Z})$. Hence, to update $\alpha_{k_{z}^{\prime}}$, we first draw $a \sim Beta(1,\alpha_{u})$. We then update $\alpha_{k_{z}^{\prime}} \leftarrow a.\alpha_{u}$ and update $\alpha_{u^{\prime}} \leftarrow (1-a).\alpha_{u}$. Similar steps are also applied to interest components.

Note that if we compare \eqref{eq:gibbsZnpUsed} to \eqref{eq:infgibbsZused}, the problem we found so far has gone since $p(z_{i}=k|\textbf{z}_{-i},\textbf{z},\textbf{t})$ will never have zero probability even if $N_{r_{i},z_{-i}} = 0$.

% will go back to add these stuffs later
%[seems to be too lazy to say this.. if have enough energy, I might do this]
At the end of each iteration, we use the same method \cite{Teh04hierarchicaldirichlet} to sample $\alpha$ and $\beta$ and update hyperparameters $\gamma_{z}$, $\gamma_{x}$, $\mu_{z}$, $\mu_{x}$ using the method described in \cite{Escobar95bayesiandensity}. We refer to this infinite version of ITM as ``Interest Topic Model with Hierarchical Dirichlet Process'' (HDPITM) for the rest of the paper.

% Computational Complexity here please...
%[Aug 30] computational complexity
For the computational complexity, although $N_{Z}$ and $N_{X}$ are both set to approach $\infty$, the computational cost of each iteration, however, does not approach $\infty$. Considering \eqref{eq:gibbsZnpUsed} and \eqref{eq:gibbsZnpNew}, sampling of $z_{i}$ only involves currently instantiated topics plus one ``collapsed topic'', which represents all currently unused topics. Similarly, the sampling of $x_{i}$ only involves currently instantiated interests plus one. For a particular iteration, a computational cost for HDP can therefore be approximated as $N_{K} \times (\kappa \times \bar{N_{Z}}+1)$. And that for HDPITM can be approximated as $N_{K} \times (\kappa \times (\bar{N_{Z}} + \bar{N_{X}}+2))$, where $\bar{N_{Z}}$ and $\bar{N_{X}}$ are respectively the average number of topics and interests in that iteration.
% I'm not sure if this is too coarse for the time complexity...

\subsection{Performance on the synthetic data}
\label{sec:nonparamsyneval}
\begin{figure}
\begin{center}
\begin{tabular}{cc}
\includegraphics[width=2.4in]{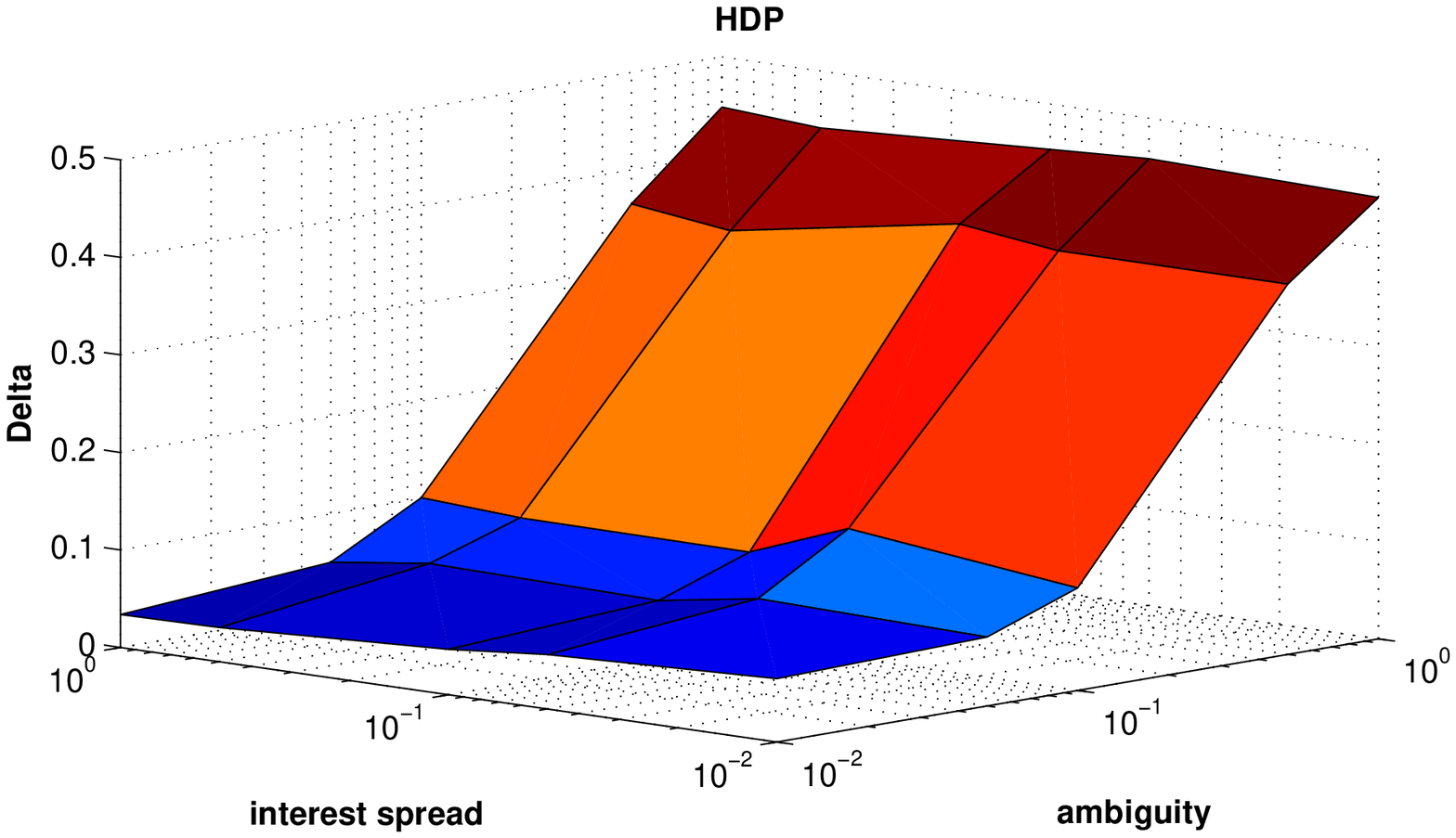}  &
\includegraphics[width=2.4in]{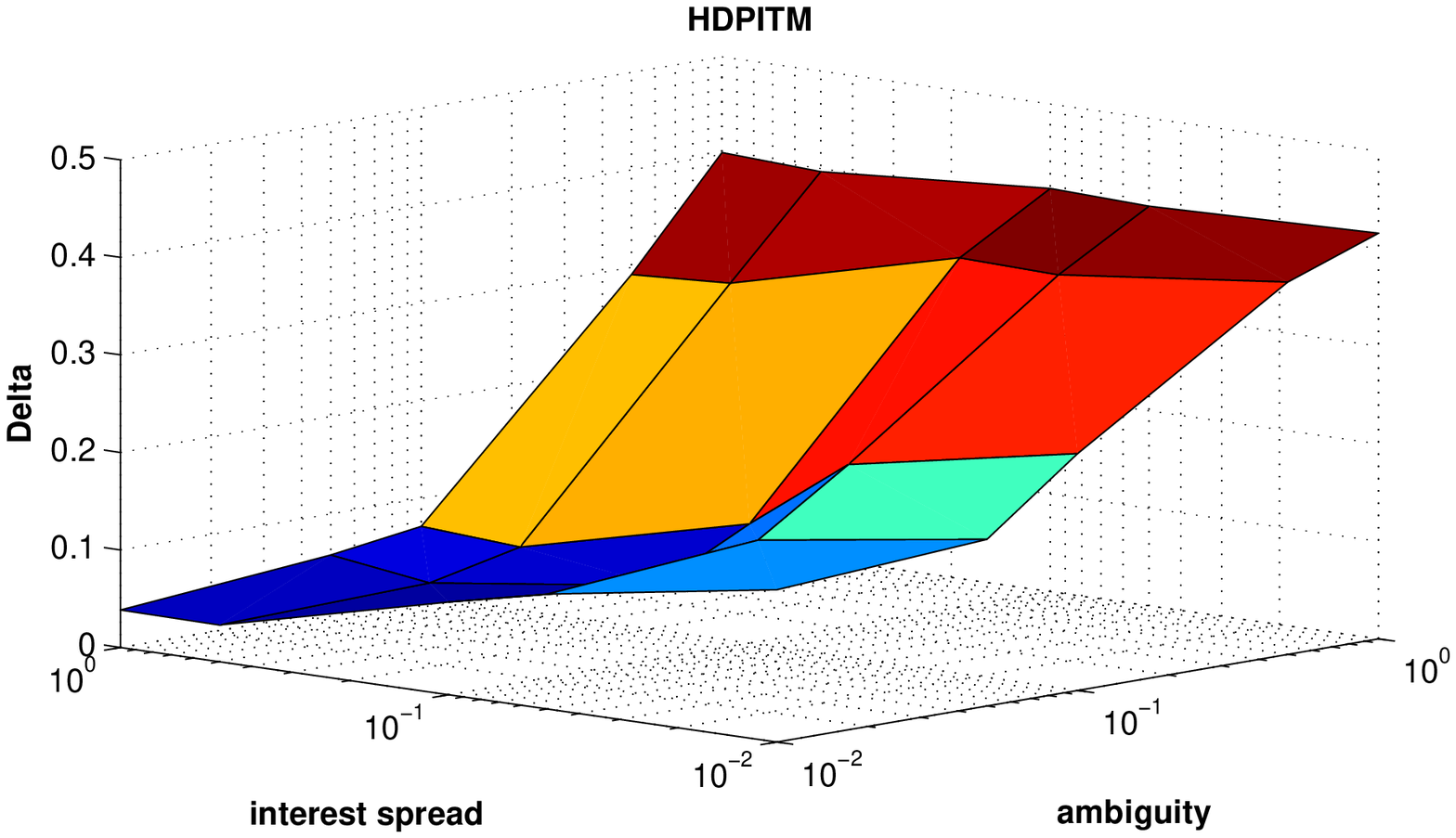} \\
(a) & (b) \\
\includegraphics[width=2.4in]{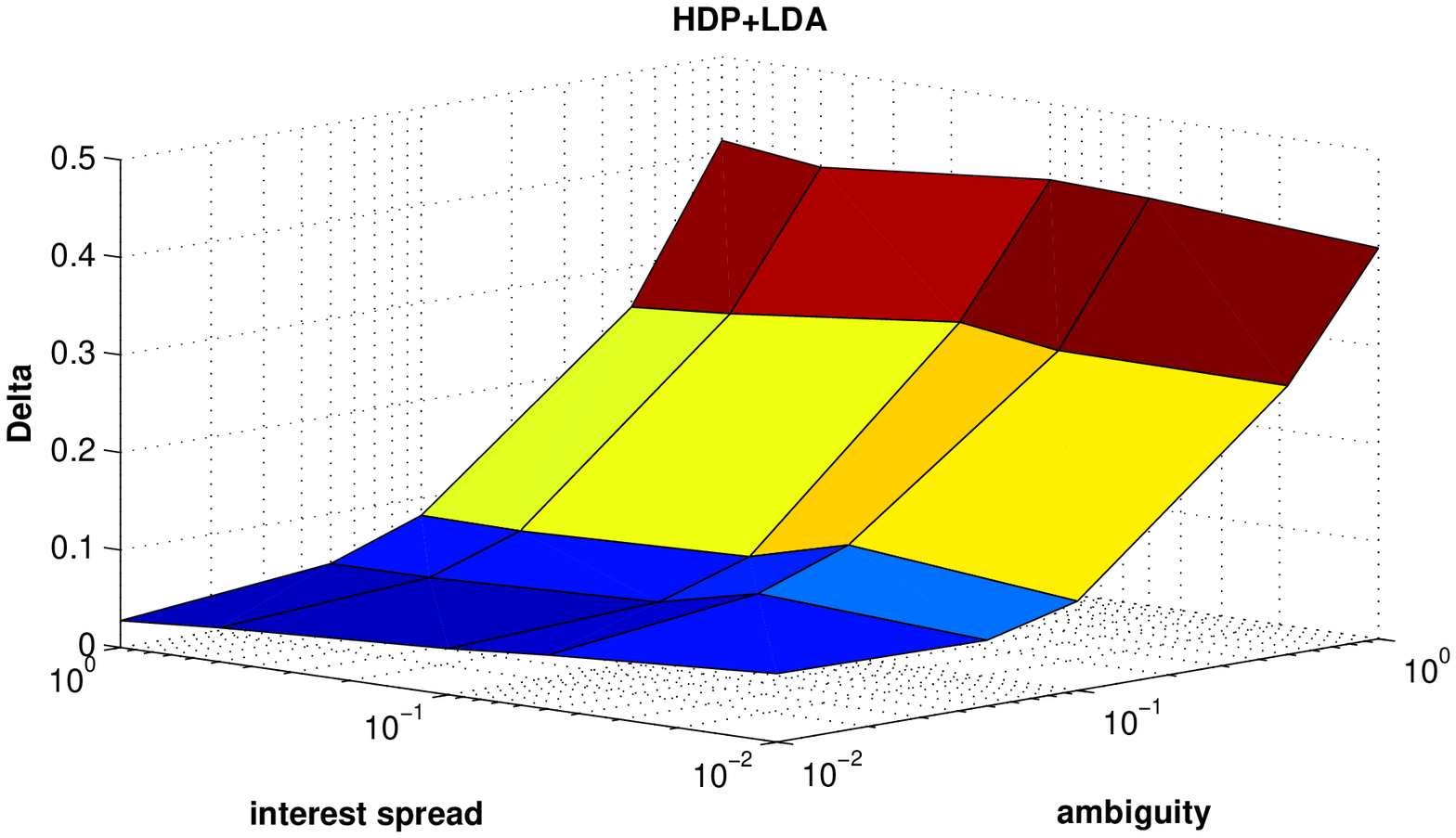}  &
\includegraphics[width=2.4in]{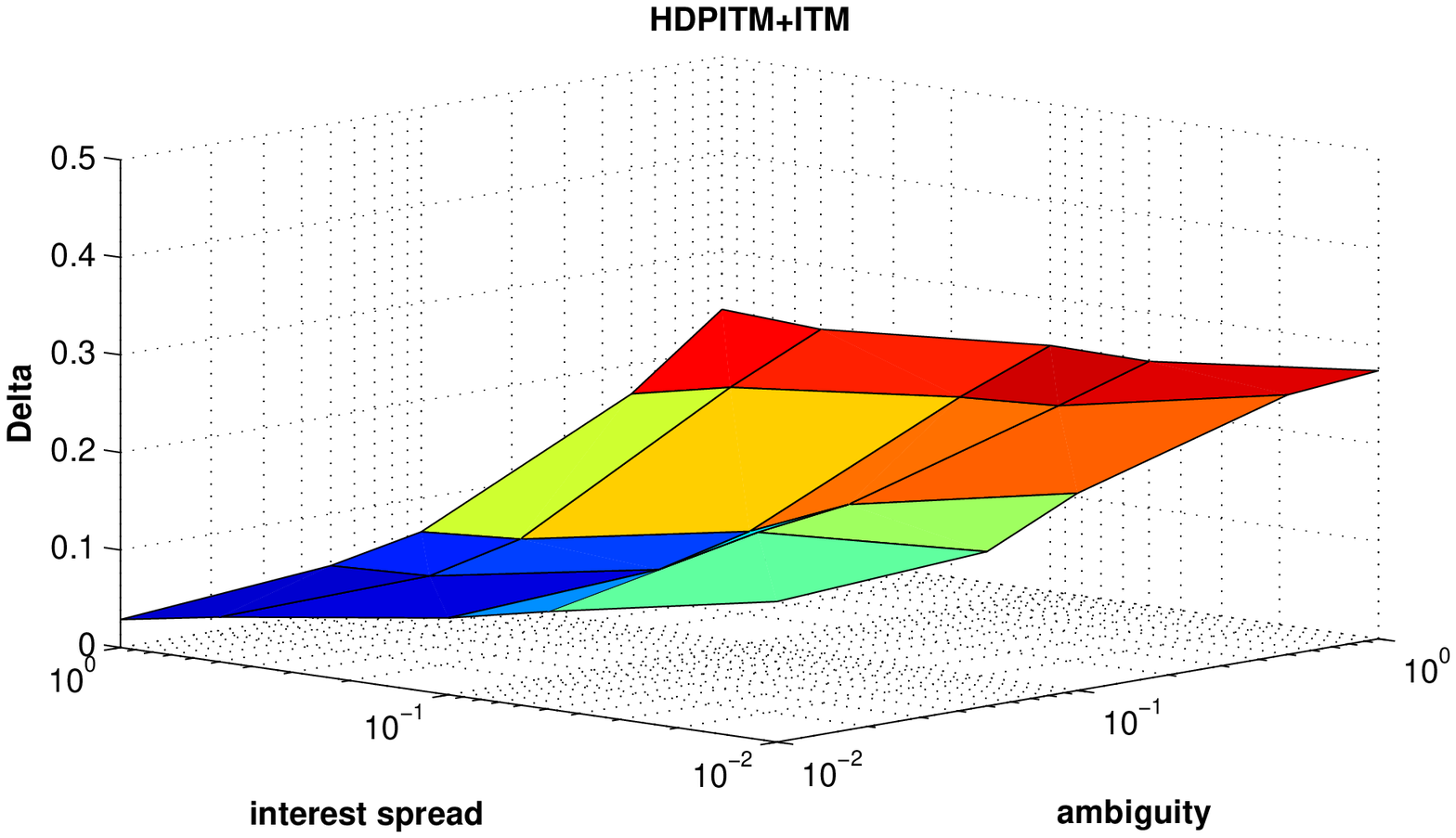} \\
(c) & (d) \\
\end{tabular}
\end{center}
\caption {This plot shows the deviation $\Delta$ between actual and learned topics on synthetic data sets, under different degrees of tag-ambiguity and user interest variation. The $\Delta$ of HDP is shown on the left (a); as that of HDPITM is on the right (b). As, (c) and (d) shows the deviation produced by HDP+LDA and HDPITM+ITM respectively. For HDP+LDA, new topics can be instantiated, and thus the number of topics can change, during the first half of the run (HDP); then all topics are freezed (no new topic can be instantiated) during the second half (LDA). And this is similar to HDPITM+ITM where we take into account user information. See \secref{sec:nonparamsyneval} for more detail.}
\label{fig:syntheticnonparam}
\end{figure}

% say something about setting for HDP , HDPITM
We ran both HDP and HDPITM to extract topic distributions, $\phi$, on the simulated data set. In each run the number of instantiated topics was initialized to ten, which equals to the actual number of topics for both HDP and HDPITM. The number of interests was initialized to three. Similar to the setting in  \secref{sec:synthetic}, topic and interest assignments were randomly initialized and then trained using 1000 iterations. Subsequently, $\phi$ was computed from the last 100 iterations.
% short report on numbers of learned topics/interests
The results are shown in \figref{fig:syntheticnonparam} (a) and (b) for HDP and HDPITM respectively. From these results, the behaviors of both model for different settings are somewhat similar to those of LDA and ITM. In particular, HDPITM can exploit user information to help disambiguate tag senses, while HDP cannot. Hence, the performance of HDPITM is better than that of HDP when tag ambiguity level is high. And since topics may lose some statistical strength under low user interest condition, HDPITM is inferior to HDP, similar to ~\figref{fig:synthetic2}(b) for the finite case.

As one can compare the plots (a) and (b) in ~\figref{fig:synthetic2} and ~\figref{fig:syntheticnonparam}, the performance of infinite model is generally worse than that of the finite one, even though we allow the former the ability to adjust topic/interest dimensions. One possible factor is that the model still allows topic/interest dimensions (configuration) to change even though the trained model is in a ``stable'' state. That would prohibit the model from optimizing its parameters for a certain configuration of topic/interest dimensions. One evidence that supports this claim is that, although the log likelihood seems to converge, the number of topics (for both models) and interests (only for HDPITM) still slightly fluctuate around a certain value.

From this speculation, we ran both HDP and HDPITM with the different strategy. In particular, we split model training into two periods. In the first period, we allow the model to adjust its configuration, i.e. the dimensions of topics and interests. In the second period, we still train the model but do not allow the dimensions of topics and interests to change. The first one is similar to the training process in the plain HDP and HDPITM. The second one is similar to that of plain LDA and ITM that use the latest configuration from the first period. In this experiment, we set the first period to 500 iterations; another 500 iterations were set for the second phase. Subsequently, $\phi$ is computed from the last 100 iterations of the second. We refer to this training strategy for HDP as HDP+LDA, and that for HDPITM as HDPITM+ITM. The overall improvement of performance using this strategy are perceived in ~\figref{fig:syntheticnonparam} (c) and (d), comparing to (a) and (b). That is, both HDP+LDA and HDPITM+ITM can produce $\phi$, which provide lower $\Delta$, under this strategy. However, HDPITM+ITM performance under the condition with low user interest and low tag ambiguity, is still inferior to HDP+LDA. This is simply because their structures are still the same to those of HDP and HDPITM respectively.

% KL [Sep 4] say something about performance differences

% discuss the behavior which is similar to finite version , but the performance is worse than the finite version.

%AP1109
% move the resource discovery task to discussion?
\subsection{Performance on the real-world data}
\label{sec:nonparamrealeval}

\begin{figure}
\begin{center}
\begin{tabular}{cc}
\includegraphics[width=2.2in]{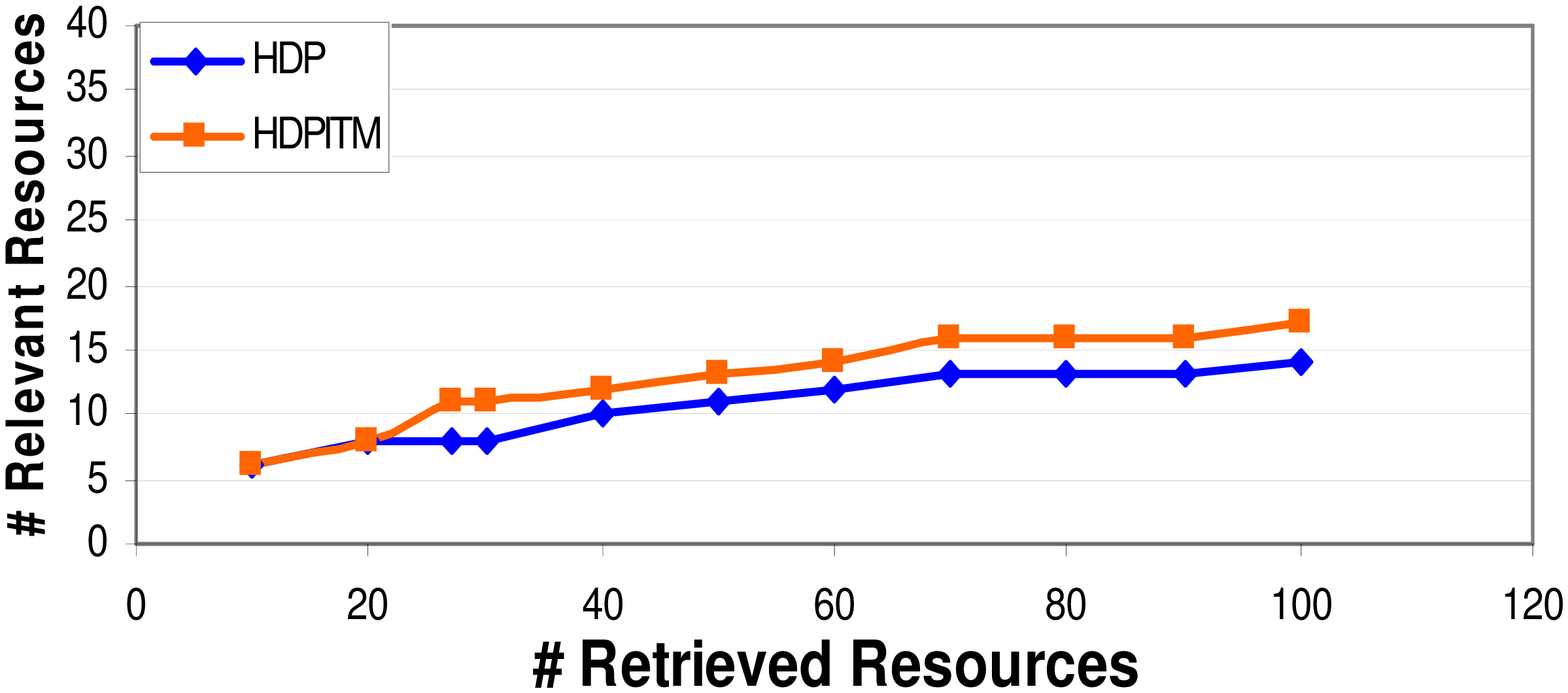} &
\includegraphics[width=2.2in]{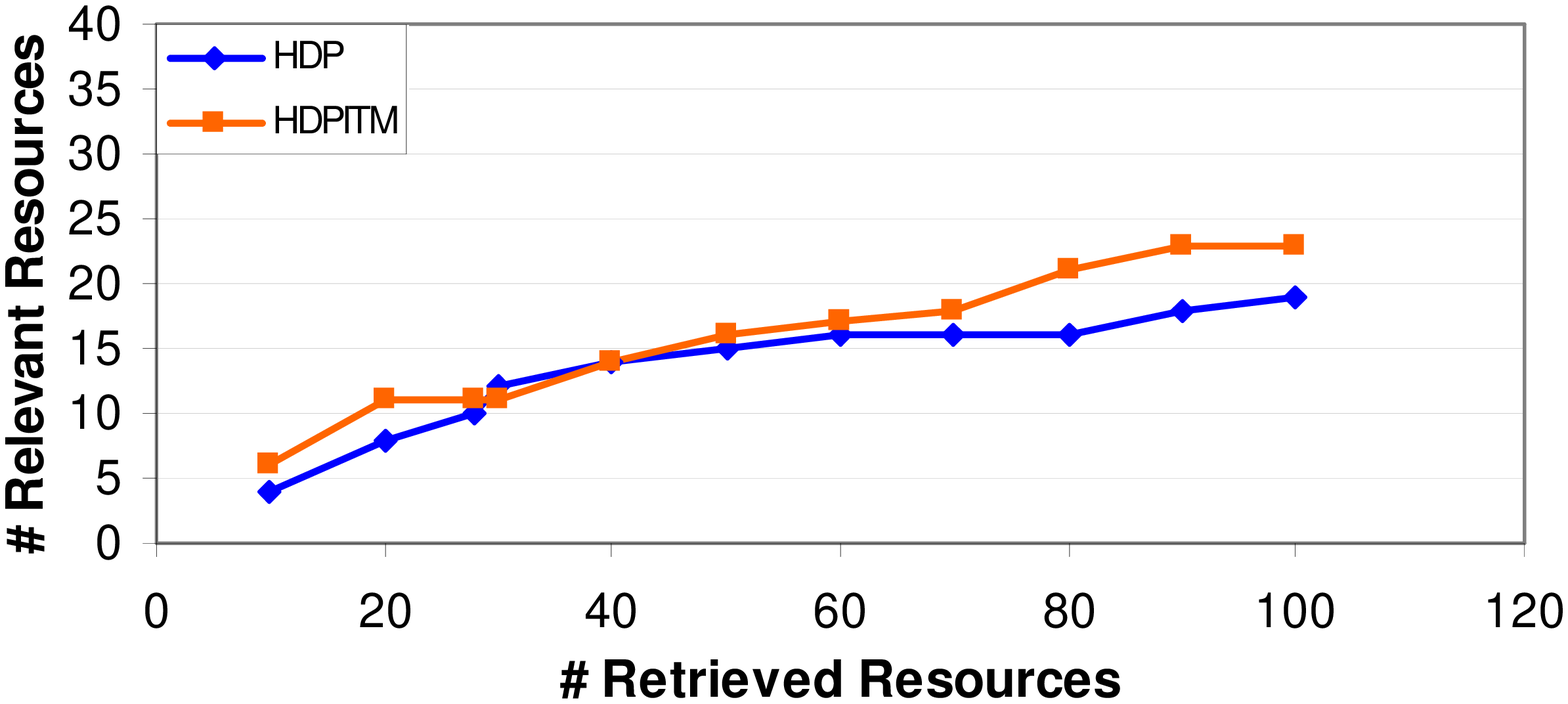} \\
(a) Flytecomm & (b) Geocoder \\
\includegraphics[width=2.2in]{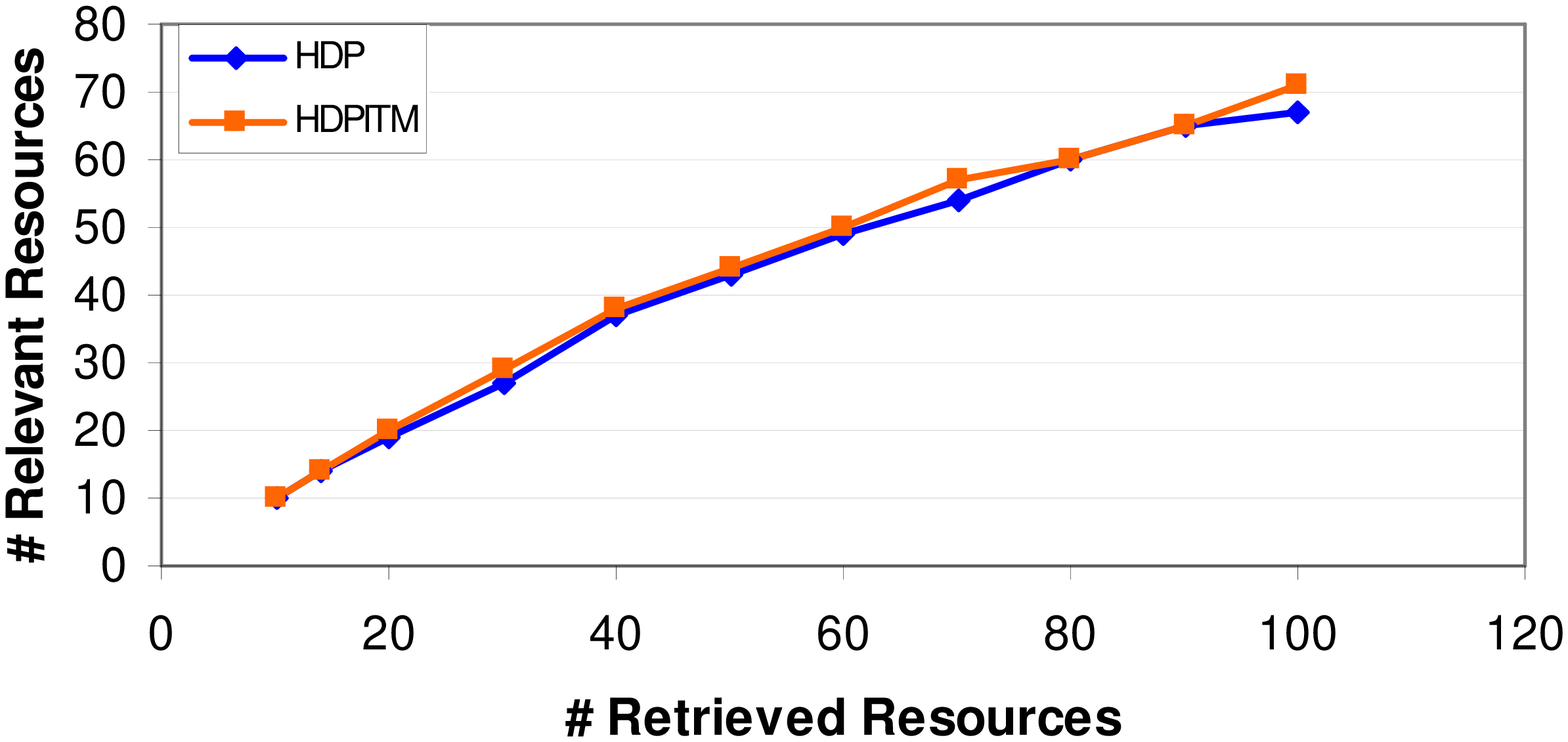} &
\includegraphics[width=2.2in]{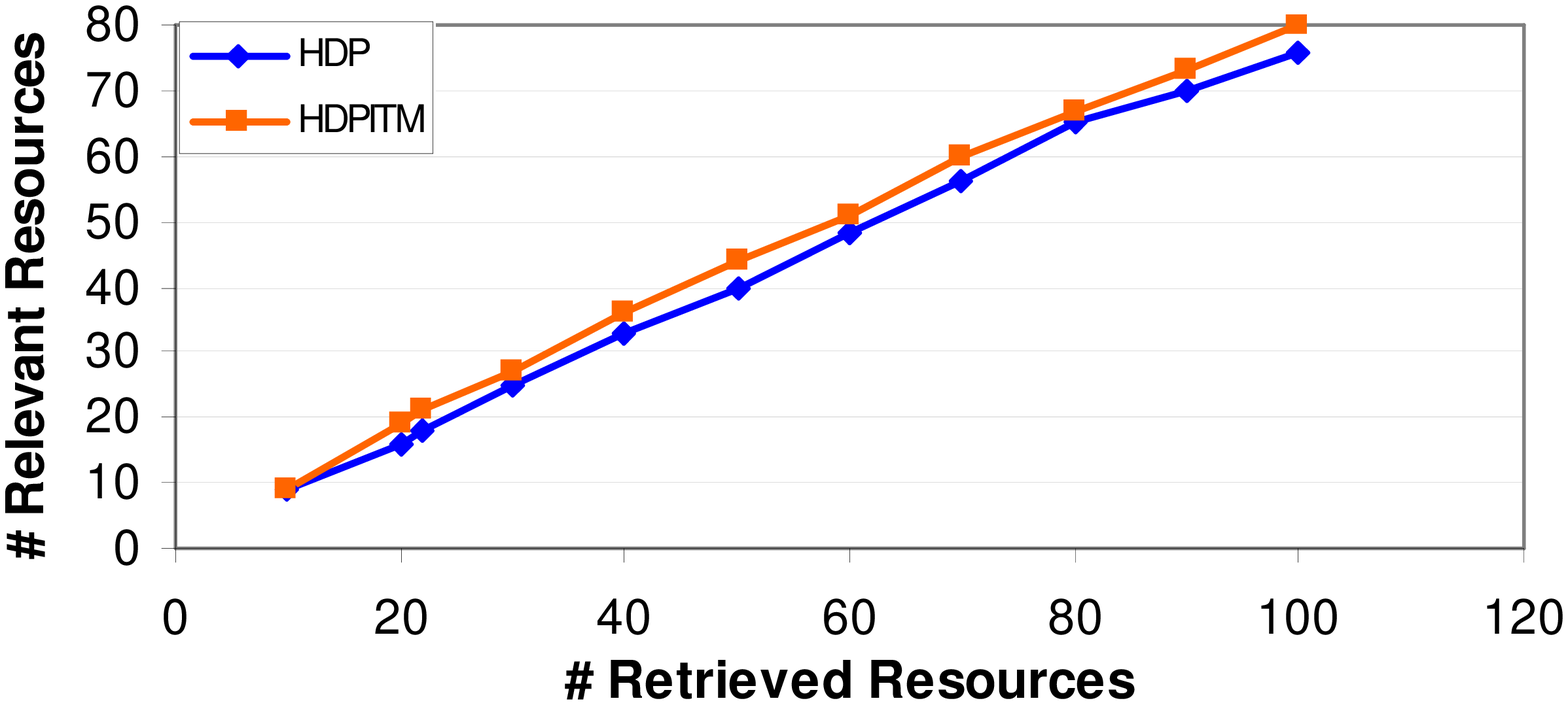}
\\
(c) Wunderground & (d) Whitepages\\
\includegraphics[width=2.2in]{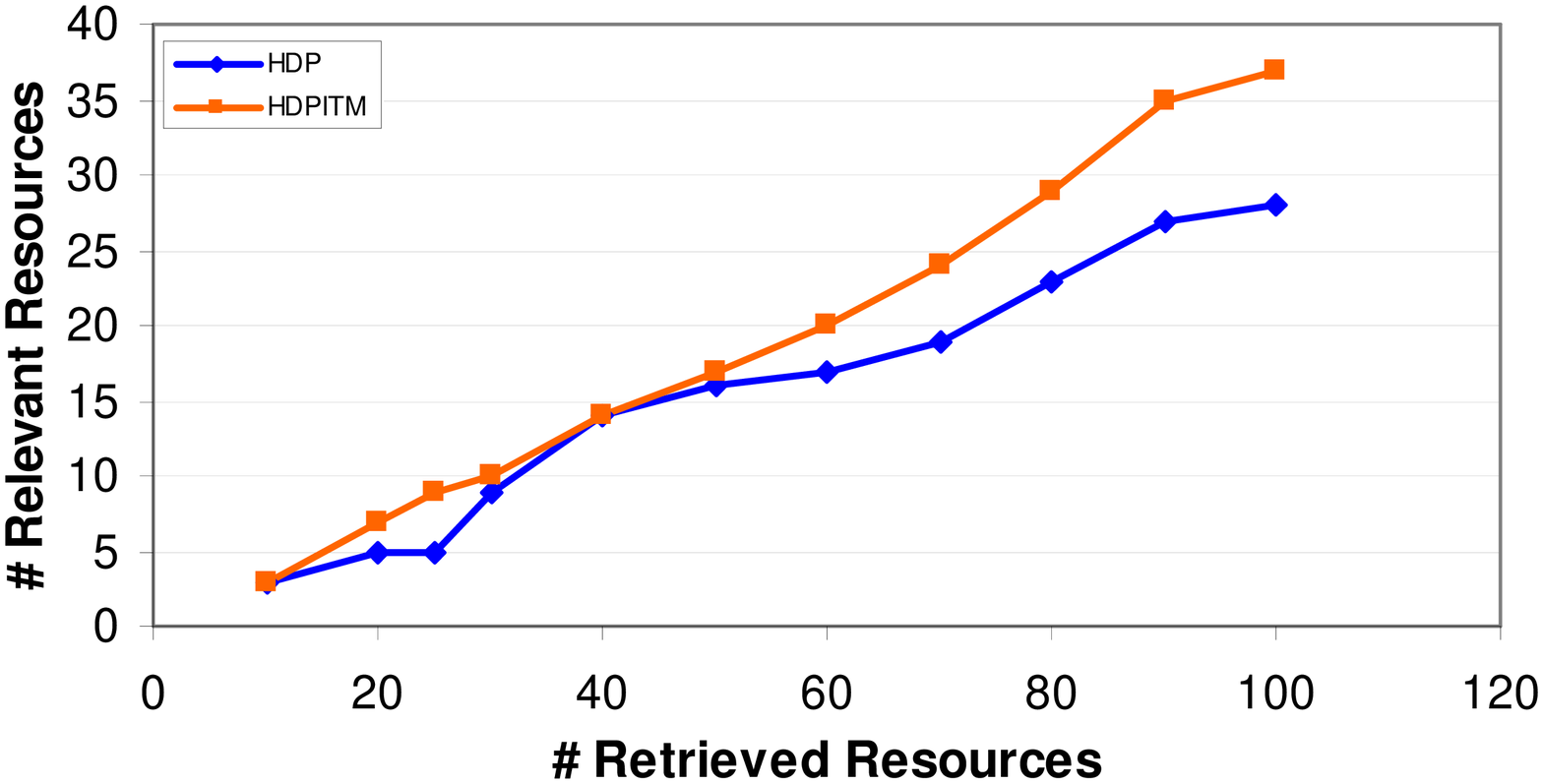} &
\includegraphics[width=2.2in]{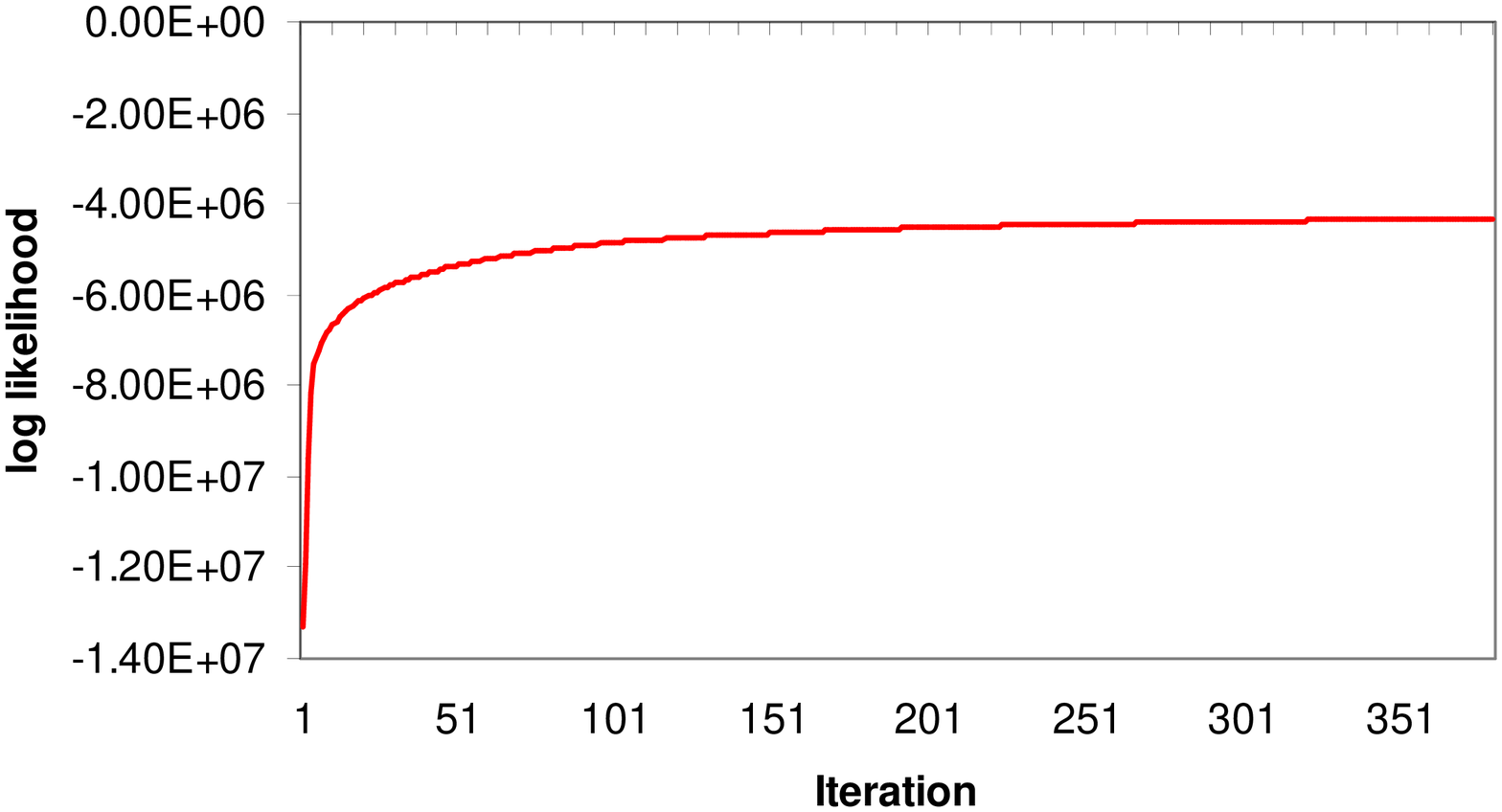}
\\
(e) Online-Reservationz & (f) Log likelood of flytecomm \comment{during parameter estimation}\\

\end{tabular}
\end{center}
% KL - 07/20/09
\caption {Performance of different methods on the five data sets (a) -- (e). Each plot shows the number of relevant resources (that are similar to the seed) within the top 100 results produced by HDP (non-parameteric version of LDA) and HDPITM (nonparametric version of ITM). Each model was initialized with 100 topics and 20 interests for HDPITM. (f) demonstrates log likelihood of the HDPITM model during parameter estimation period of \emph{flytecomm} data set. Similar behavior of the plot (f) is found in both HDP and HDPITM for all data sets.}
\label{fig:rankresults}
\end{figure}

In the experiments, we initialize the numbers of topics and interests to 100 and 20 (the number of interests is only applicable to HDPITM), and train the models on the same real-world data sets we used in \secref{sec:realitm}.  The topic and interest assignments are randomly initialized, and then both models are trained with the minimum 400 and maximum 600 iterations. For the first 100 iterations, we allow both models to instantiate a new topic or interest as required, under the constraint that the number of topics and interests does not exceed 400 and 80 respectively. If the model violates this constraint, it will exit this phase early. For the remainder of iterations, we do not allow the model to add new topics or interests (but these numbers can shrink if some topics/interests collapsed during this phase). Then, if the change in log likelihood, averaged over the 10 preceding iterations,  is less than $2\%$, the training process will enter to final learning phase. (See \figref{fig:rankresults} (f) for an example of log likelihood during training iterations.) In fact, we found that the process enters the final phase early in all data sets. In the final phase, consisting of 100 iterations, we use the topic and interest assignments in each iteration to compute the distributions of resources over topics.

The reason we limit the maximum numbers of topics, interests, and iterations over which these models are allowed to instantiate a new topic/interest, is that the numbers of users and tags in our data sets are large, and many new topics and interests could be instantiated. This would require many more iterations to converge, and the models would require more memory than is available on the desktop machine we used in the experiments.\footnote{At maximum, we can only allocate memory for 1,300 Mbytes.} We would rather allow the model to ``explore'' the underlying structure of data within the constraints --- in other words, find a configuration which is best suited to the data under a limited exploration period and then fit the data within that configuration.
At the end of the parameter estimation, the numbers of allocated topics of HDP models for \emph{flytecomm}, \emph{geocoder}, \emph{wunderground}, \emph{whitepages} and \emph{online-reservationz} was $171$, $174$, $197$, $187$ and $175$ respectively. The numbers of allocated topics and interests in HDPITM are $\langle307,43\rangle$, $\langle329,44\rangle$, $\langle231,81\rangle$, $\langle225,78\rangle$ and $\langle207,72\rangle$ respectively, which is bigger than those inferred by HDP in all cases. These results suggests that user information allows the HDPITM discover more detailed structure.

%result charts

% say a little bit about the results
HDPITM performs somewhat better than HDP in \emph{flytecomm}, \emph{online-reservationz}, and \emph{geocoder} data sets. Its performance for \emph{wunderground} and \emph{whitepages}, however, is almost identical to HDP. As in \secref{sec:realitm}, this is possibly due to high interest variation among users. We suspect that weather and directory services are of interest to all users, and are therefore bookmarked by a large variety of users.

\section{Related Research}
\label{sec:previous}
Modeling social annotation is an emerging new field, but it has intellectual roots in two other fields: document modeling and collaborative filtering. It is relevant to the former in that one can view a resource being annotated by users with a set of tags to be analogous to a document, which is composed of words from the document's authors. Usually, the numbers of users involved in creating a document is much less than those involved in annotating a resource. In regard to collaborative rating systems, annotations created by users in a social annotation system are analogous to object ratings in a recommendation system. However, users only provide one rating to the object in a recommendation system, but they usually annotate an object with several keywords. Therefore, there are several relevant threads of research connecting our work to earlier ones in these areas.

In relation to document modeling, our work is conceptually motivated by the Author-Topic model (AT)~\cite{TopicModelSmyth2004}, where we can view a user who annotate a resource as an author who composes a document. In particular, the model explains the process of document generation, governed by author profiles, in forms of distributions of authors over topics. However, this work is not directly applicable to social annotations.
This is because, first, in social annotation context, we know who generates a tag on a certain resource; therefore, the author selection process in AT, which selects one of co-authors to be responsible for a generation of a certain document word, is not needed in our context. Second, co-occurrences of user-tag pairs for a certain bookmark are very sparse, i.e., there are fewer than 10 tags per bookmark. Thus, we need to group users who share the same interests together to avoid the sparseness problem. Third, AT has no direct way to estimate distributions of resources over topics since there are only author-topic and topic-word associations in the model. One possible indirect way is to compute this from an average over all distributions of authors over topics\comment{, of the authors who actually bookmarked that resource}. Our model, instead, explicitly models this distribution, and since it uses profiles of groups of similar users, rather than those of an individual, the distributions are expected to be less biased.

Several recent works apply document modeling to a social annotation. One study~\cite{ChineseModel} applies the multi-way aspect model~\cite{HofmannPLSA2001,Popescul01} to social annotations on \emph{Delicious}. The model does not explicitly separate user interests and resource topics as our model does, and thus cannot exploit user variations to learn better distributions of resources over topics, as we showed in \cite{delicious07::iiweb}.

\cite{ZhouUCA08} introduced a generative model of the process of Web page creation and annotation. The model, called User Content Annotator (UCA), includes words found in Web documents, in addition to tags generated by users to annotate these documents. The authors explore this model in the context of improving IR performance. In this work, a bag of words (tags and content) is generated from two different sources --- the document creator and annotator. Although UCA takes documents' contents into account, unlike our model, it makes several assumptions, which we believe do not hold for real-world data. The first assumption is that annotators conceptually agree with the original document's authors (and therefore, share the the same topic space), whereas ITM relaxes this assumption. The second assumption is that users and documents have the same types of distribution over topics, whereas ITM separates \emph{interests} from \emph{topics}. In fact, without documents' content, UCA is almost identical to the Author Topic model \cite{TopicModelSmyth2004}, except for the fact that owners tags are explicitly known, and thus, it shares AT's drawbacks. Another technical drawback of UCA is the following:  if a particular tagged Web document has no words (e.g., a Web service, Flickr photo, or YouTube video), UCA would then take into account the taggers only, and lose the variable $d$ that represents the document. Further computation is required to infer $p(z|d)$, the probability of a topic given a document, which is required for the content discovery task we are investigating.

% collaborative filtering
% mention about Jin et. al. works.
Collaborative filtering  was one of the first successful social applications.  Collaborative filtering is a technology used by recommender systems to find users with similar interests by asking them to rate items. It then compares their ratings to find users with similar opinions, and recommends to users new items that similar users liked. Among of recent works in collaborative filtering area, ~\cite{Jin_mixmodel_clfr06} is most relevant to ours. In particular, the work describe a mixture model for collaborative filtering that takes into account users' intrinsic preferences about items. In this model, item rating is generated from both the item type and user's individual preference for that type. Intuitively, like-minded users would have similar rating on the same item types (e.g., movie genres). When predicting a rating of a certain item for a certain user, the user's previous ratings on other items will be used to infer a like-minded group of users. Then, the ``common'' rating on that item from the users of that group is the prediction. This collaborative rating process is very similar to that of collaborative tagging. The only technical difference is that each ``item'' can have multiple ``ratings'' (in our case, tags) from a single user. This is because an item usually has multiple subjects and each subject can be represented using multiple terms.

There exist, however, major differences between ~\cite{Jin_mixmodel_clfr06} and our work. We use the probabilistic model to discover a ``resource description'' despite users annotating resources with potentially ambiguous tags. Our goal is not to predict how a user will tag a resource (analogous to predicting a rating user will give to an item), or discovering like-minded groups of users, which our algorithm could also do. The main purpose of our work is to recover the actual ``resource description'' from noisy observations generated by different users. In essence, we hypothesize that there is actual description of a certain resource and users select and then annotate the resource with that description partially according to their ``interest'' or ``expertise''. In this work, we also demonstrate that when taking into account individual difference in the process, the inferred resource descriptions are not biased toward individual variation as much as those that do not take this issue into account. Another technical difference is that the model is not implemented in fully Bayesian, and uses point estimation to estimate its parameters, which is criticized to be susceptible to local maxima \cite{GriffithsTopic04,TopicModel06}. Moreover, it can not be extended to allow numbers of topics/interests to be flexble as ours; thus, the strong assumption on the number of topics and interests is required.

Rather than modeling social annotation, \cite{EfficientBrowsing07} concentrates on an approach that helps users efficiently navigate the Social Web.
Although the work share some similar challenges, e.g., tag ambiguity, with ours, the solution proposed in that work is rather different. In particular, the work exploits user activity to resolve ambiguity -- as a user selects more tags, the topic scope gets more focused. Consequently, the recently suggested tags associate with fewer and fewer senses, helping to disambiguate the tag. Our approach does not rely on such user activity to disambiguate tag senses; instead, we exploit user interests to do this, since tag sense is correlated with a group of users who share interests. On an applications level, this approach and ours are also different. In particular, the former approach is suitable for situations when users activity and labeled data is available, and can be exploited to filter information on the fly. Our approach, on the other hand, utilizes social annotation only. It is more suitable for batch jobs without user's intervention; for example, the automatic resource discovery task for mashup applications~\cite{AmbiteISWC09}.

\section{Conclusion}
\label{sec:conclusion}
We have presented a probabilistic model of social annotation that takes into account the users who are creating the annotations. We argued that our model is able to learn a more accurate topic description of a corpus of annotated resources by exploiting individual variations in user interests  and vocabulary to help disambiguate tags. Our experimental results on collections of annotated Web resources from the social bookmarking site \emph{Delicious} show that our model can effectively exploit social annotation on the resource discovery task.

One issue that our model does not address is tag bias, probably caused by expressiveness of users with high interests in a certain domain. In general, a few users use many more tags than others in annotating resources. This will bias the model toward these users' annotations, causing the learned topic distributions to deviate from the actual distributions. One possible way to compensate for this is to tie the number of tags to individual interests in the model. ITM also does not at present allow us to include other sources of evidence about documents, e.g., their contents. It would be interesting to extend ITM to include content words, which will make this model more attractive for Information Retrieval tasks.

Since our model is more computationally expensive than other models that ignore user information, e.g. LDA,  it is not practical to blindly apply our approach to all data sets. Specifically, our model cannot exploit individual variation in the data that has low tag ambiguity and small individual variation, as shown in \secref{sec:synthetic}. In this case, our model can only produce small improvement or even similar performance to that of the simpler models. For a practical reason, a heuristic for determining level of tag ambiguity and user variation would be very beneficial in order to determine if the complex model is preferable to the simpler one. Ratios between a number of tags and that of users or that of resources may provide some clues.

As we model the social annotation process by taking into account all essential entities; namely, users, resources and tags, we can apply the model to other applications. For example, one can straightforwardly apply the model to personalize search~\cite{ChineseModel,Lerman07flickrsearch}. It can also be used to suggest tags to a user annotating a new resource, in the same spirit as rating predictions in Collaborative Filtering.

\section*{Appendix}
\label{sec:gibbsderive}
%[Aug 27]
We begin to derive Gibbs sampling equations for ITM in \secref{sec:finiteitm} from the joint probability of $\textbf{t}$, $\textbf{x}$ and $\textbf{z}$ of all tuples. Suppose that we have $n$ tuples. Their joint probability is defined as

\begin{eqnarray}
p&(&t_{i},x_{i},z_{i} ; i=1:n) \nonumber \\ &=& \int p(t_{i},x_{i},z_{i}|\psi,\phi,\theta;i=1:n).p(\psi, \phi,\theta)d\langle\psi,\phi,\theta\rangle \nonumber \\
&=& c\cdot\int \prod_{i=1:n} {( \psi_{u_{i},x_{i}}\cdot\phi_{r_{i},z_{i}}\cdot\theta_{t_{i},z_{i},x_{i}}  )}\cdot\prod_{u,x}{\psi_{u,x}^{\beta/N_{X}-1}} \nonumber \\ &\cdot& \prod_{r,z}{\phi_{r,z}^{\alpha/N_{Z}-1}}\cdot\prod_{t,x,z}{\theta_{t,x,z}^{\eta/N_{T}-1}} d\langle\psi,\phi,\theta\rangle \nonumber \\
&=& c\cdot\int{\prod_{u,x}{\psi_{u,x}^{\sum_{i}\delta_{u}(x_{i},x)+\beta/N_{X}-1}} d(\psi)}\cdot\int{\prod_{r,z}{\phi_{r,z}^{\sum_{i}\delta_{r}(z_{i},z)+\alpha/N_{Z}-1}} d(\phi)} \nonumber \\ &\cdot&\int{\prod_{t,z,x}{\theta_{t,z,x}^{\sum_{i}\delta_{z,x}(t_{i},t)+\eta/N_{T}-1}} d(\theta)} \nonumber \\
&=& c\cdot\prod_{r}{(\frac{\prod_{z}{\Gamma(\sum_{i}{\delta_{r}(z_{i},z)}+\alpha/N_{Z})}}{\Gamma(N_{r}+\alpha)})}
\cdot\prod_{u}{(\frac{\prod_{x}{\Gamma(\sum_{i}{\delta_{u}(x_{i},x)}+\alpha/N_{X})}}{\Gamma(N_{u}+\beta)})} \nonumber \\
&\cdot& \prod_{z,x}{(\frac{\prod_{z,x}{\Gamma(\sum_{i}{\delta_{z,x}(t_{i},t)}+\eta/N_{T})}}{\Gamma(N_{z,x}+\eta)})}
\label{eq:bitm1}
\end{eqnarray}
\noindent where c = $(\frac{\Gamma(\alpha)}{\Gamma(\alpha/N_{Z})^{z}})^{r}\cdot(\frac{\Gamma(\beta)}{\Gamma(\beta/N_{X})^{x}})^{u}\cdot(\frac{\Gamma(\eta)}{\Gamma(\eta/N_{T})^{t}})^{(z,x)}$ and $\delta_{r}(z_{i},z)$ is a function which returns 1 if $z_{i} = z$ and $r_{i} = r$ otherwise 0. $N_{r}$  represents a number of all tuples associated with resource $r$. Similarly, $N_{x,z}$ represents a number of all tuples associated with interest $x$ and topic $z$.

By rearranging ~\eqref{eq:bitm1}, we obtain
\begin{eqnarray}
p(t_{i},x_{i},z_{i} ; i=1:n) &=& \prod_{r}{(\frac{\Gamma(\alpha)}{\Gamma(N_{r}+\alpha)})}\cdot\prod_{r,z}{(\frac{\Gamma(\sum_{i}{\delta_{r}(z_{i},z)+\alpha/N_{Z}})}{\Gamma(\alpha/N_{Z})})} \nonumber \\
&\cdot& \prod_{u}{(\frac{\Gamma(\beta)}{\Gamma(N_{u}+\beta)})}\cdot\prod_{u,x}{(\frac{\Gamma(\sum_{i}{\delta_{u}(x_{i},x)+\beta/N_{X}})}{\Gamma(\beta/N_{X})})} \nonumber \\
&\cdot& \prod_{x,z}{(\frac{\Gamma(\eta)}{\Gamma(N_{x,z}+\eta)})}\cdot\prod_{x,z,t}{(\frac{\Gamma(\sum_{i}{\delta_{x,z}(t_{i},t)+\eta/N_{T}})}{\Gamma(\eta/N_{T})})}
\label{eq:bitm2}
\end{eqnarray}

Suppose that we have a new tuple and we index this tuple with $k$ (say $k = n+1$ for convenience). From \eqref{eq:bitm2}, we can derive a joint probability of this new tuple $k$ and all other previous tuples as follows

\begin{eqnarray}
p&(&t_{k},x_{k},z_{k},t_{i},x_{i},z_{i} ; i=1:n) \nonumber \\
&=& \frac{\Gamma(\alpha)}{\Gamma(N_{r=r_{k}}+\alpha+1)}\cdot(\prod_{r \neq r_{k}}{\frac{\Gamma(\alpha)}{\Gamma(N_{r}+\alpha)}}) \cdot
\frac{\Gamma(\sum_{i}{\delta_{r=r_{k}}(z_{i},z_{k})}+\alpha/N_{Z}+1)}{\Gamma(\alpha/N_{Z})} \nonumber \\
&\cdot&(\prod_{r \neq r_{k}, z \neq z_{k}}{\frac{\Gamma(\sum_{i}{\delta_{r}(z_{i},z)}+\alpha/N_{Z})}{\Gamma(\alpha/N_{Z})}})
\cdot \frac{\Gamma(\beta)}{\Gamma(N_{u=u_{k}}+\beta+1)}\cdot(\prod_{u \neq u_{k}}{\frac{\Gamma(\beta)}{\Gamma(N_{u}+\beta)}}) \nonumber \\
&\cdot&
\frac{\Gamma(\sum_{i}{\delta_{u=u_{k}}(x_{i},x_{k})}+\beta/N_{X}+1)}{\Gamma(\beta/N_{X})}
\cdot(\prod_{u \neq u_{k}, x \neq x_{k}}{\frac{\Gamma(\sum_{i}{\delta_{u}(x_{i},x)}+\beta/N_{X})}{\Gamma(\beta/N_{X})}}) \nonumber \\
&\cdot&
\frac{\Gamma(\eta)}{\Gamma(N_{x=x_{k},z=z_{k}}+\eta+1)}\cdot(\prod_{x \neq x_{k},z \neq z_{k}}{\frac{\Gamma(\eta)}{\Gamma(N_{x,z}+\eta)}}) \nonumber \\
&\cdot&
\frac{\Gamma(\sum_{i}{\delta_{x=x_{k},z=z_{k}}(t_{i},t_{k})}+\eta/N_{T}+1)}{\Gamma(\eta/N_{T})}
\cdot (\prod_{x \neq x_{k}, z \neq z_{k}, t \neq t_{k}}{\frac{\Gamma(\sum_{i}{\delta_{x,z}(t_{i},t)}+\eta/N_{T})}{\Gamma(\eta/N_{T})}}) \nonumber \\
\end{eqnarray}

For the tuple $k$, suppose that we only know the values of $x_{k}$ and $t_{k}$ while that of $z_{k}$ is unknown. The joint probability of all tuples, excluding $z_{k}$ is as follows.

% following derivation can be omitted
\begin{eqnarray}
p&(&t_{k},x_{k},t_{i},x_{i},z_{i} ; i=1:n) \nonumber \\
&=& \frac{\Gamma(\alpha)}{\Gamma(N_{r=r_{k}}+\alpha)}\cdot(\prod_{r \neq r_{k}}{\frac{\Gamma(\alpha)}{\Gamma(N_{r}+\alpha)}})
\cdot
\frac{\Gamma(\sum_{i}{\delta_{r=r_{k}}(z_{i},z)}+\alpha/N_{Z})}{\Gamma(\alpha/N_{Z})} \nonumber \\
&\cdot&(\prod_{r \neq r_{k}, z \neq z_{k}}{\frac{\Gamma(\sum_{i}{\delta_{r}(z_{i},z)}+\alpha/N_{Z})}{\Gamma(\alpha/N_{Z})}})
\cdot
\frac{\Gamma(\beta)}{\Gamma(N_{u=u_{k}}+\beta+1)}\cdot(\prod_{u \neq u_{k}}{\frac{\Gamma(\beta)}{\Gamma(N_{u}+\beta)}}) \nonumber \\
&\cdot&
\frac{\Gamma(\sum_{i}{\delta_{u=u_{k}}(x_{i},x_{k})}+\beta/N_{X}+1)}{\Gamma(\beta/N_{X})}
\cdot(\prod_{u \neq u_{k}, x \neq x_{k}}{\frac{\Gamma(\sum_{i}{\delta_{u}(x_{i},x)}+\beta/N_{X})}{\Gamma(\beta/N_{X})}}) \nonumber \\
&\cdot&
\frac{\Gamma(\eta)}{\Gamma(N_{x=x_{k},z=z_{k}}+\eta)}\cdot(\prod_{x \neq x_{k},z \neq z_{k}}{\frac{\Gamma(\eta)}{\Gamma(N_{x,z}+\eta)}})
\cdot
\frac{\Gamma(\sum_{i}{\delta_{x=x_{k},z=z_{k}}(t_{i},t_{k})}+\eta/N_{T})}{\Gamma(\eta/N_{T})} \nonumber \\
&\cdot&(\prod_{x \neq x_{k}, z \neq z_{k}, t \neq t_{k}}{\frac{\Gamma(\sum_{i}{\delta_{x,z}(t_{i},t)}+\eta/N_{T})}{\Gamma(\eta/N_{T})}}) \nonumber \\
\label{eq:bitm3}
\end{eqnarray}

By dividing \eqref{eq:bitm2} by \eqref{eq:bitm3}, we can obtain the posterior probability of $z_{k}$ given all other variables as follows

% \Gamma(X) = (X-1)! if X is an integer
\begin{eqnarray}
p&(&z_{k}|t_{k},x_{k},t_{i},x_{i},z_{i} ; i=1:n) \nonumber \\
&=&
\frac{\Gamma(N_{r=r_{k}}+\alpha)}{\Gamma(N_{r=r_{k}}+\alpha+1)}\cdot\frac{\Gamma(\sum_{i}{\delta_{r=r_{k}}(z_{i},z)+\alpha/N_{Z}+1})}{\Gamma(\sum_{i}{\delta_{r=r_{k}}(z_{i},z)+\alpha/N_{Z}})} \nonumber \\
&\cdot& \frac{\Gamma(N_{x=x_{k},z=z_{k}}+\eta)}{\Gamma(N_{x=x_{k},z=z_{k}}+\eta+1)}\cdot\frac{\Gamma(\sum_{i}{\delta_{x=x_{k},z=z_{k}}(t_{i},t_{k})+\eta/N_{T}+1})}{\Gamma(\sum_{i}{\delta_{x=x_{k},z=z_{k}}(t_{i},t_{k})+\eta/N_{T}})} \nonumber \\
&=&
\frac{\sum_{i}{\delta_{r=r_{k}}(z_{i},z_{k})}+\alpha/N_{Z}}{N_{r=r_{k}}+\alpha}.\frac{\sum_{i}{\delta_{x=x_{k},z=z_{k}}(t_{i},t_{k})}+\eta/N_{T}}{N_{x=x_{k},z=z_{k}}+\eta} \nonumber \\
&=&
\frac{N_{r=r_{k},z=z_{k}}+\alpha/N_{Z}}{N_{r=r_{k}}+\alpha}\cdot\frac{N_{x=x_{k},z=z_{k},t=t_{k}}+\eta/N_{T}}{N_{x=x_{k},z=z_{k}}+\eta}
\label{eq:posteriorz}
\end{eqnarray}

Intuitively, we can perceive from \eqref{eq:posteriorz} that $\frac{N_{r=r_{k},z=z_{k}}+\alpha/N_{Z}}{N_{r=r_{k}}+\alpha}$ tell us how resource $r$ is likely to be described by the topic $z$; as the later part, $\frac{N_{x=x_{k},z=z_{k},t=t_{k}}+\eta/N_{T}}{N_{x=x_{k},z=z_{k}}+\eta}$ tell us how tag $t$ is likely to be chosen given interest $x$ and $z$.

Similarly, we can obtain the posterior probability of $x_{k}$ as we did for $z_{k}$.
\begin{eqnarray}
p(x_{k}|t_{k},z_{k},t_{i},x_{i},z_{i} ; i=1:n) &=&
\frac{N_{u=u_{k},x=x_{k}}+\beta/N_{X}}{N_{u=u_{k}}+\beta} \nonumber \\
&\cdot&\frac{N_{x=x_{k},z=z_{k},t=t_{k}}+\eta/N_{T}}{N_{x=x_{k},z=z_{k}}+\eta}
\label{eq:posteriorx}
\end{eqnarray}

We can now generalize \eqref{eq:posteriorz} and \eqref{eq:posteriorx} for sampling posterior probabilities of topic $z$ and interest $x$ of a present tuple $i$ given all other tuples. We define $N_{r_{i},z_{-i}}$ as the number of all toples having $r = r_{i}$ and $z$ but excluding the present tuple $i$. Similarly, $N_{z_{-i},x_{i},t_{i}}$ is a number of all tuples having $x = x_{i}$, $t = t_{i}$ and $z$ but excluding the present tuple $i$. As $\textbf{z}_{-i}$ represents all topic assignments except that of the tuple $i$.

\begin{eqnarray}
p(z_{i}|\textbf{z}_{-i},\textbf{x},\textbf{t}) = \frac{N_{r_{i},z_{-i}}+\alpha/N_{Z}}{N_{r_{i}}+\alpha-1}.\frac{N_{z_{-i},x_{i},t_{i}}+\eta/N_{T}}{N_{z_{-i},x_{i}}+\eta}
\label{eq:gibbsZa}
\end{eqnarray}

\begin{eqnarray}
p(x_{i}|\textbf{x}_{-i},\textbf{z},\textbf{t}) = \frac{N_{u_{i},x_{-i}}+\beta/N_{X}}{N_{u_{i}}+\beta-1}\cdot\frac{N_{x_{-i},z_{i},t_{i}}+\eta/N_{T}}{N_{x_{-i},z_{i}}+\eta}
\label{eq:gibbsXa}
\end{eqnarray}

\begin{acks}
We would like to thank anonymous reviewers for providing useful comments and suggestions to improve the manuscript. This material is based in part upon work supported by the National Science Foundation under Grant Numbers CMMI-0753124 and IIS-0812677. Any opinions, findings, and conclusions or recommendations expressed in this material are those of the authors and do not necessarily reflect the views of the National Science Foundation.
\end{acks}

\bibliographystyle{acmtrans}

\end{document}